\documentclass[10pt,twocolumn,letterpaper]{article}

\usepackage[pagenumbers]{cvpr} 

\usepackage{graphicx}
\usepackage{amsmath}
\usepackage{amssymb}
\usepackage{booktabs}
\usepackage{xcolor, colortbl}
\usepackage[accsupp]{axessibility}

\definecolor{LightGrey}{rgb}{0.9,0.9,0.9}

\usepackage[pagebackref,breaklinks,colorlinks]{hyperref}

\usepackage[capitalize]{cleveref}
\crefname{section}{Sec.}{Secs.}
\Crefname{section}{Section}{Sections}
\Crefname{table}{Table}{Tables}
\crefname{table}{Tab.}{Tabs.}

\usepackage{times}
\usepackage{epsfig}

\usepackage{latexsym}
\usepackage{float}
\usepackage{makecell}
\usepackage{comment}
\usepackage{booktabs}
\usepackage{multirow}

\renewcommand\cellgape{\Gape[4pt]}
\usepackage{amsfonts}
\usepackage{bbm}
\usepackage{multicol}
\usepackage[font=small,labelfont=bf,tableposition=top]{caption}

\usepackage{xcolor}

\newcommand{\ra}{\rightarrow}

\def\mypar#1{\vspace{1mm}{\noindent\bf #1}\hspace{1mm}}

\begin{document}

\title{FlexIT: Towards Flexible Semantic Image Translation}

\def \ours {{FlexIT}\xspace}

\def\cvprPaperID{8356} 
\def\confName{CVPR}
\def\confYear{2022}

\author{Guillaume Couairon\\
Facebook AI Research\\

{\tt\small gcouairon@fb.com}
\and
Asya Grechka \\
Meero \\
{\tt\small 	asya.grechka@meero.com}
\and
Jakob Verbeek \\
Facebook AI Research\\
{\tt\small jjverbeek@fb.com}
\and
Holger Schwenk \\
Facebook AI Research\\
{\tt\small schwenk@fb.com}
\and
Matthieu Cord \\
LIP6 \\
{\tt\small matthieu.cord@lip6.fr}
}

\twocolumn[{%
\renewcommand\twocolumn[1][]{#1}%
\maketitle
\begin{center}
\vspace{-1em}
    \centering
    \captionsetup{type=figure}
    \includegraphics[width=.80\textwidth]{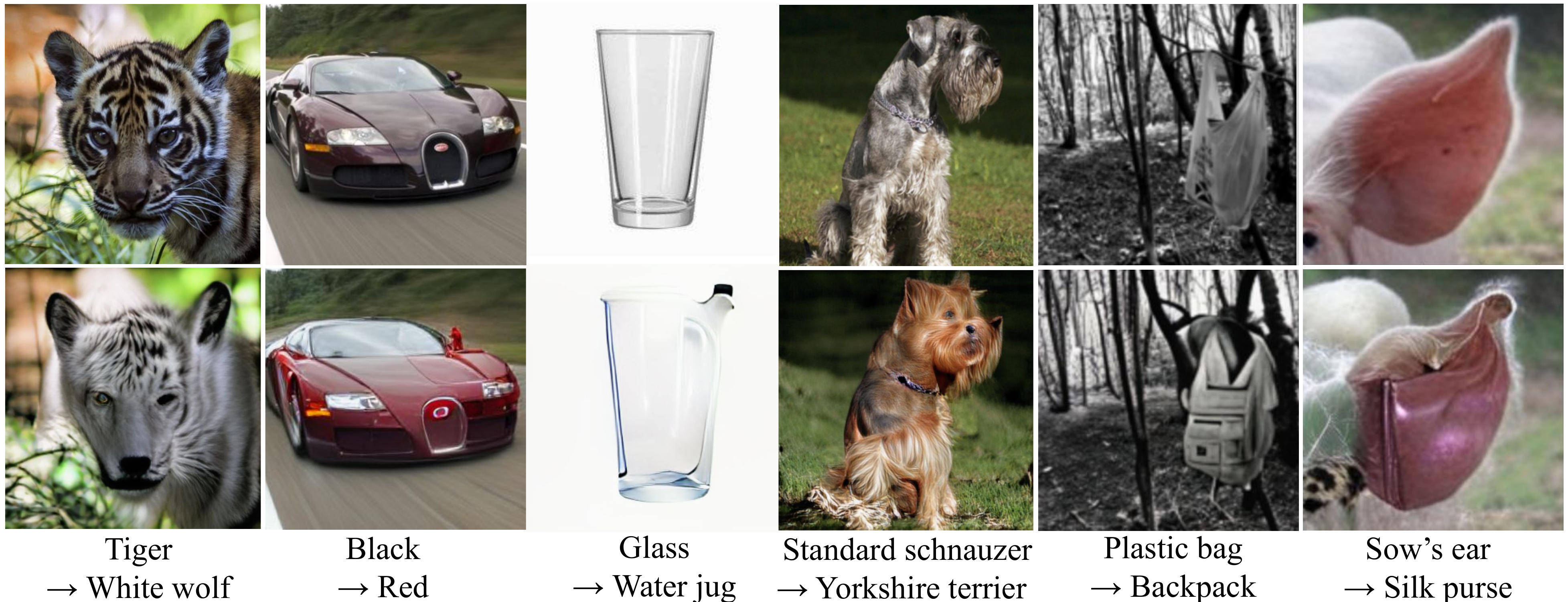}
    \captionof{figure}{\ours transformation examples. From top to bottom: input image, transformed image, and text query.
    }
    \label{fig:main_demo}
\end{center}%
}]

\maketitle
\begin{abstract}
\vspace*{-1em}
Deep generative models, like GANs, have considerably improved the state of the art in image synthesis, and are  able to generate near photo-realistic images in structured domains such as human faces. Based on this success, recent work on image editing proceeds by projecting images to the GAN latent space and manipulating the latent vector. However, these approaches are limited in that only images from a narrow domain can be transformed, and  with only a limited number of editing operations. We propose FlexIT, a novel method which can take any input image and a user-defined text instruction for editing. Our method achieves flexible and natural editing, pushing the limits of semantic image translation. First, FlexIT combines the input image and text into a single target point in the CLIP multimodal embedding space. Via the latent space of an  autoencoder, we iteratively transform the input image toward the target point, ensuring coherence and quality with a variety of novel regularization terms. We propose an evaluation protocol for semantic image translation, and  thoroughly evaluate our method on  ImageNet. Code will be made publicly available.
\end{abstract}

\section{Introduction}

The old saying goes: \emph{``You can’t make a silk purse from a sow’s ear.''} Or can you?
Truly flexible and powerful semantic image editing is elusive, and current work is  limited in terms of possible input images and edit operations. Research in deep generative image models has seen significant progress in recent years, with GANs in particular  generating near photo-realistic samples in domains such as human and animal faces~\cite{karras20nips} or object-centric images~\cite{brock19iclr}. Moreover, recent ``style-based'' GANs, like StyleGAN~\cite{karras19cvpr, karras20cvpr, Karras2021stylegan3}, have an impressively disentangled latent space, where  performing copy-pastes between two latent vectors transfers the corresponding styles in the image space. 

Consequently, significant research efforts have been put into using pre-trained GANs for semantic image edition. Through specific latent-space manipulation, high-level attributes such as age or gender can be identified and edited in a realistic manner~\cite{shen2020, abdal2020styleflow, zhuang2021enjoy, harkonen2020ganspace}. These approaches, however, present several caveats. 
First, contrary to generated latents, inferred latent codes representing  real images have been shown to react poorly to latent editing operations~\cite{grechka2021magec}.
Although recent methods~\cite{grechka2021magec, tov2021designing, zhu2020indomain} improve editability, input images are still highly limited to the distribution of the generative network. Moreover, edit operations are also limited to the semantics identified in the latent space via a pre-trained classifier~\cite{zhuang2021enjoy, shen2020, abdal2020styleflow} or through a semi-automatic manner~\cite{voynov20icml, harkonen2020ganspace}. These semantics are specific to the single domain the GAN was trained on, such as age or apparent gender in the case of faces.
Some flexibility \wrt the input images can be obtained by training a GAN to directly modify the images, known as image-to-image translation. These methods learn a transformation between two domains, using paired data~\cite{isola17cvpr, pix2pixhd, gaugan} or unpaired data~\cite{zhu17iccv, choi2020starganv2}. However, these models only learn a  single transformation, or combinations thereof~\cite{wang20ijcv}, specific to the training data,  limiting the scope of their applicability.

We  tackle these challenges with a unified framework which modifies an input image based on a user-defined text query of the form $(S \ra T)$, like \textit{cat} → \textit{dog}. For this \textit{semantic image translation} task, the goal is to make minimal image modifications  while transforming the image as requested.
We leverage CLIP\cite{radford21clip},  which combines text and image representations in one powerful multimodal embedding space. This space is used to define our target point, based on the embeddings from the user input. We perform a per-image optimization procedure, using specific  strategies to ensure image quality and relevance to the transformation query. Our method requires only fixed pre-trained components, and can thus be used off-the-shelf  without  requiring any training. The image is optimized in the latent space of an auto-encoder, rather than a GAN, which greatly enlarges the scope of possible input images. This allows for truly flexible image edits; as Fig.~\ref{fig:main_demo} shows, even a sow's ear can be  changed into a silk purse. 

We   propose an evaluation protocol for the task of semantic image translation.
Evaluation is based on three criteria: (i) the transformed image should correctly correspond to the text query, (ii) the output image should look natural,  and (iii) visual elements irrelevant to the text query should remain unchanged. 
We thoroughly evaluate our model on ImageNet, and  demonstrate quantitatively and qualitatively the superiority of our method against baselines, broadening the horizon of text-driven image editing.

\section{Related Work}

\mypar{Image editing.} Deep generative networks, like GANs, have given rise to numerous image editing applications, ranging from photography retouching~\cite{shi2020benchmark}, image inpainting~\cite{yu18cvpr}, object insertion~\cite{gafni20cvpr}, 
domain translation~\cite{zhu17iccv,yu2019multi}, colorization~\cite{isola17cvpr}, super-resolution~\cite{johnson16eccv,ledig17cvpr},
among many others. 
Automatic user-driven image editing aims at providing the user control to modify an image, by tweaking segmentation masks~\cite{ling2021editgan}, scene graphs~\cite{dhamo2020semantic}, or class labels~\cite{casanova21nips}. 
Allowing the user to provide unstructured  free-form text queries  is more challenging. 
Close to our objective,  ManiGAN~\cite{li2020manigan} aims at performing text-based edits by training a model to refine the details of an image based on its textual description. Their quantitative evaluation protocol uses transformation queries on the COCO dataset by considering unaligned (image, caption) pairs, resulting in possibly incoherent transformation queries.
We carefully design our evaluation protocol to  avoid such cases.

\mypar{Image latent space.}
While GANs are highly effective as generative models, inference of the latent variable given an image is intractable.
Even though joint learning of an inference network has been proposed, see \eg ~\cite{donahue17iclr,dumoulin17iclr}, the mode-seeking training dynamics of GANs are not suited for good reconstruction performance beyond the training distribution (or even within it, if modes are dropped).
Variational autoencoders~\cite{kingma14iclr}, on the other hand,  offer an inference network by construction, and their likelihood-based training objective ensures accurate reconstructions.

Vector-quantized  variational autoencoders (VQ-VAE)~\cite{oord17nips,razavi19nips}, which discretize the latent space, have been found to offer both good reconstructions as well as compelling samples. 
In particular, VQ-GAN~\cite{esser2021taming,yu21arxiv} further improves reconstructions by  including an adversarial loss term to train the autoencoder.
In our work,  we adopt the VQ-GAN autoencoder,  and edit  images in its latent space.

\begin{figure*}
    \centering
    \vspace{-1em}
    \includegraphics[width=\linewidth]{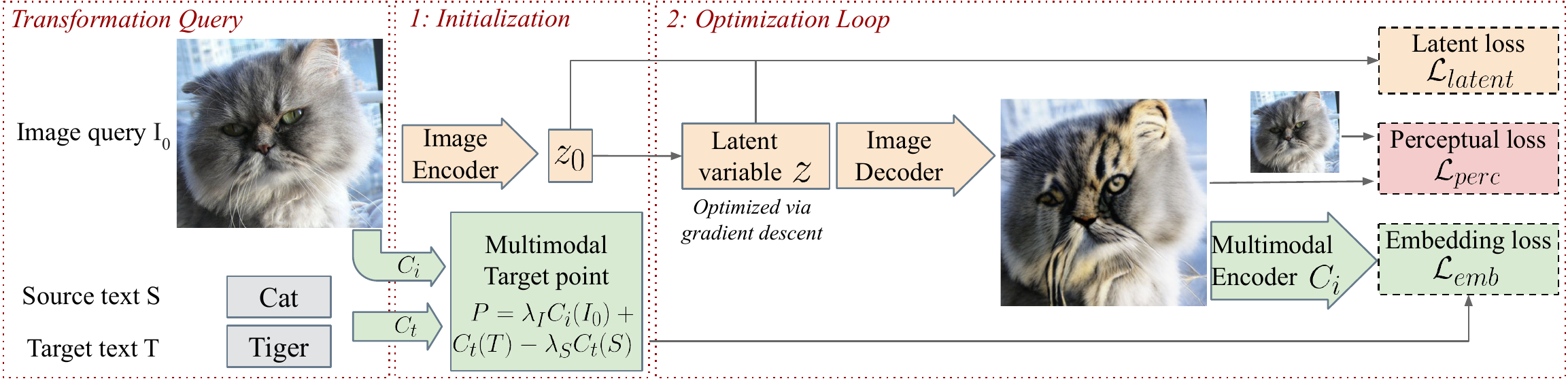}
    \caption{FlexIT optimization framework: components  involving the multimodal latent space  colored in green; those involving the image latent space in yellow; those involving the LPIPS distance in pink. Given a transformation query $(I_0, S, T)$, we first compute a target point $P$ in the multimodal embedding space, and we encode $I_0$ in the image latent space to get $z_0$. Then, for a fixed number of steps, we update the latent variable $z$ (initialized with $z_0$) to get closer to the target point $P$. 
    We  add two regularization terms: the LPIPS perceptual distance between the input image and the output image, and a latent distance between $z$ and $z_0$. All networks are frozen, only $z$ is updated.}
    \label{fig:method}
    \vspace{-1em}
\end{figure*}

\mypar{Latent space manipulation.}
The introduction of ``style-based" GANs, such as  StyleGAN~\cite{karras19cvpr, karras20cvpr, Karras2021stylegan3} significantly improved the disentanglement of the latent space, leading to a surge of research into its interpretation and manipulation. By using an auxiliary classifier, a simple approach consists in finding linear boundaries in the latent space separating binary attributes~\cite{shen2020, zhuang2021enjoy, Goetschalckxganalyze}, which allows to edit attributes by ``walking'' in the orthogonal latent direction.
StyleFlow~\cite{abdal2020styleflow} proposes a non-linear approach by learning the latent transformations using normalizing flows. Other methods~\cite{harkonen2020ganspace, voynov20icml} operate without a pre-trained classifier and find the transformations in an unsupervised manner, requiring a manual labelling process to interpret and annotate the ``discovered'' transformations. Rather such  restricted sets of possible edit dimensions, we target more general transformations described by free-text.

\mypar{Semantic alignment with CLIP.}
To align images and text, CLIP~\cite{radford21clip} learns encoders that map both modalities to a shared latent space in which they can easily be compared and combined. Vision encoders are based on ResNets \cite{he2016deep} and Vision Transformers \cite{dosovitskiy2020image}.

CLIP, trained on 400M web-crawled image/text pairs with a simple contrastive  InfoNCE loss\cite{oord18arxiv},  can provide a robust differentiable signal for image synthesis and editing, used in conjunction with diffusion models~\cite{kim2021diffusionclip}, and vector strokes generators \cite{frans2021clipdraw}. 
Similarly to us, StyleCLIP~\cite{patashnik2021styleclip} transforms images based on text queries via alignment in CLIP's latent space. However it relies on the latent space of StyleGAN2 to optimize the image, which requires training a separate generative and latent space inference model per application domain.

\section{FlexIT framework for semantic editing}

An overview of our image transformation approach is depicted in Figure~\ref{fig:method}. 
It relies on three pre-trained components. 
First, we edit the input image in a  latent space, with the requirement that a wide range of  images can be encoded and decoded back to an RGB image with minimal distortion. We chose the VQGAN autoencoder~\cite{esser2021taming} for that purpose.
Second, we embed the text query and input image in a multimodal embedding space, to define the optimization target for the modified image. We use the CLIP~\cite{radford21clip} multimodal embedding spaces.
Finally, to ensure that the modified image remains similar to the input, we control its distance to the input image with the LPIPS perceptual distance \cite{zhang18cvpr} computed with a VGG \cite{simonyan15iclr} backbone.

\mypar{Optimization scheme.}
The core idea of the FlexIT method is to edit the input image in a latent space, guided by a high-level semantic objective defined in the multimodal embedding space. Let $E$ be the image encoder, $D$ the image decoder and $(C_t, C_i)$ the multimodal encoders for text and image respectively. Given an input image $I_0$ and a textual transformation $S \rightarrow T$, we first initialize \ours by computing the initial latent image representation as $z_0 = E(I_0)$ and the target multimodal point $P$ as
\begin{equation} 
P = C_t(T) + \lambda_{I} C_i(I_0) - \lambda_S C_t(S). \label{eq:p}
\end{equation}
We choose to use a multimodal embedding space since it allows text and image modalities to be combined together in a meaningful way: semantic transformations defined by textual embeddings can be applied to images with linear operations~\cite{jia2021scaling}. In this context, our target point $P$ can be seen as an image embedding that has been semantically modified with textual embeddings, by removing the source class information ($-\lambda_S E_t(S)$) and adding the target class information ($+E_t(T)$). Since we don't know what is the optimal linear combination of image and text embeddings, we consider $\lambda_I$ and $\lambda_S$ as parameters which will be validated on our development set.

To find an output image which, when encoded in the multimodal embedding space, gets as close as possible to the target point, we optimize the embedding loss: 
\begin{equation}
    \mathcal{L}_{emb}(z) = \Vert C_i(D(z)) - P \Vert_2^2.
\end{equation}

We add two regularization terms to the embedding loss, to encourage that only the content related to the transformation query is changed. 
Without regularization, the optimization scheme can alter any part of the image if this helps in getting closer to the multimodal target point, which we have found to yield unnatural artifacts. 
The distance to the input image $I_0$ is controlled with a LPIPS distance:
\begin{equation}
\mathcal{L}_{perc}(z) = d_{LPIPS} (D(z), I_0). 
\end{equation}

To enforce staying in parts of the latent space that are well decoded by our image decoder, we use a regularization term with respect to the initial latent code $z_0$. 
We use a $\ell_2$ norm at each spatial position $i$ of the latent code, and sum these norms across  spatial positions to obtain the loss: 
\begin{equation}
\mathcal{L}_{latent}(z) = \sum_i \Vert z^i - z_0^i \Vert_2.
\end{equation}
This $\ell_{2,1}$ loss encourages sparse $z^i$ changes, \textit{i.e.}\ limiting changes in spatial locations, which is aligned with our objective to transform a localized part of the input image.

Finally, note that  $\lambda_I$ in Eq.\ (\ref{eq:p}) also acts as a regularization parameter, by encouraging  the input  and output image to be close in the multi-modal embedding space.

The total loss we optimize can be written as:
\begin{equation}
\mathcal{L}_{total}(z) = \mathcal{L}_{emb}(z) + \lambda_p \mathcal{L}_{perc}(z)  + \lambda_z \mathcal{L}_{latent}(z).
\end{equation}
After initialization, 
the latent image variable $z$ is updated via gradient descent with a fixed learning rate $\mu$ for a fixed number of steps $N$, while keeping all network weights frozen. Following the implementation of the Fast Gradient Method \cite{dong2018boosting}, we normalize the gradient before the update.

\begin{figure}
    \centering
    \includegraphics[width=\linewidth]{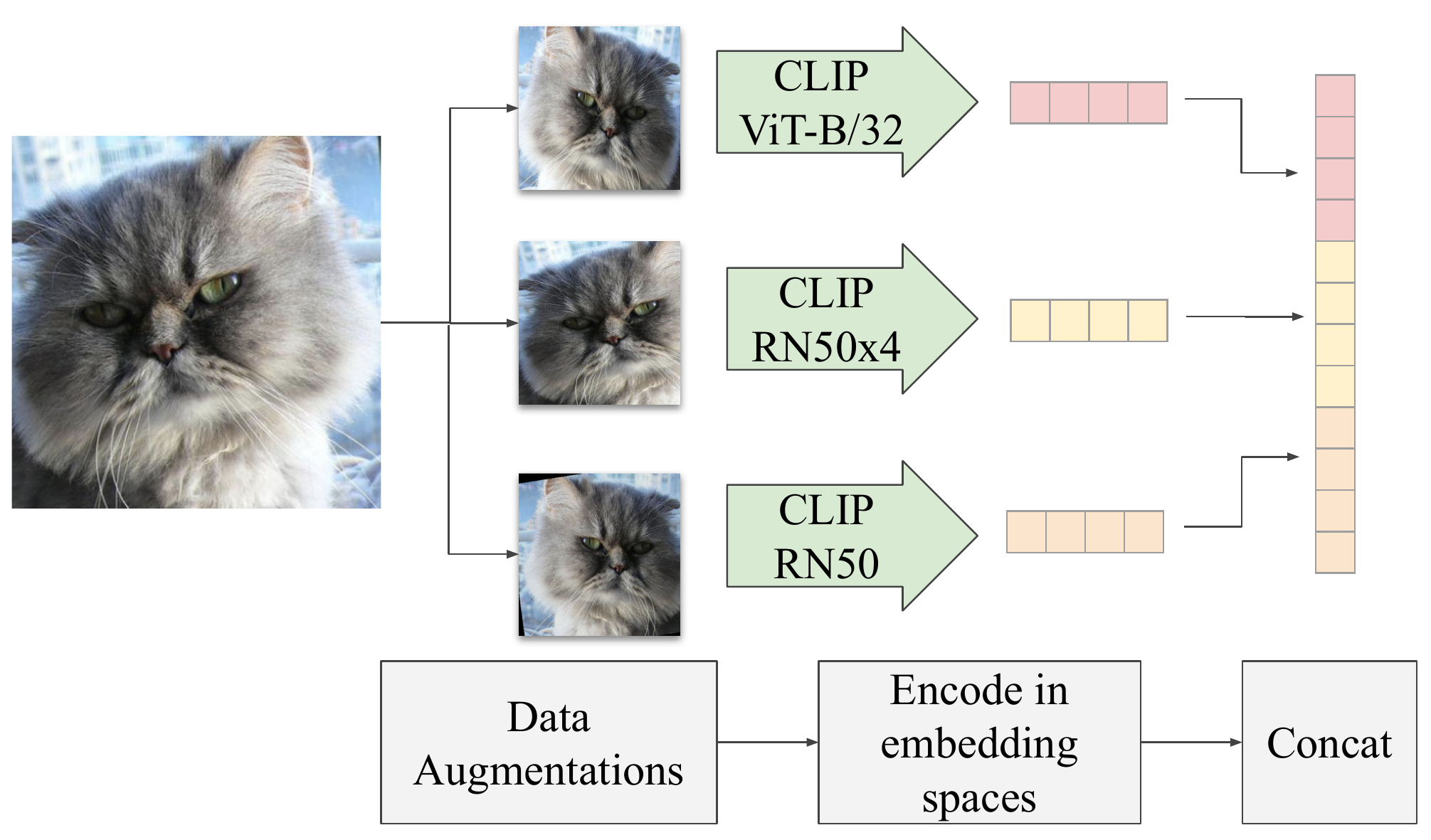}
    \caption{
    Architecture of our robust CLIP-based image encoder, which combines three different  encoders by concatenation.}
    \label{fig:clips}
\end{figure}

\mypar{Image optimization space.}
The distance to the multi-modal target point is a differentiable loss that can be optimized via gradient descent. A straightforward approach consists in performing gradient descent directly in the pixel-space. However, this type of image representation lacks a  prior on low-level image statistics. 
By optimizing over a  latent variable instead, the  image is obtained as the output of a neural-network based decoder. Choosing an autoencoder, like that of VQGAN, lets us (i) make use of the decoder's low-level priors, which guides the optimization problem towards images that exhibit at least low-level consistency; and (ii) encode and decode images in its latent space with little distortion. The spatial dimensions in the VQGAN latent space allows to edit specific parts of the image independently, contrary to GANs which typically rely on more global latent variables. 
Although GANs generate realistic images with stronger priors, it is problematic to optimize their latent space for two reasons: first, GANs work  well on narrow distributions (such as human faces), but do not work as well when trained  on a much wider distribution;
second, even with a GAN trained on a wide distribution such as that of ImageNet, it is hard to faithfully reconstruct  an image using its  latent space.

We report on experiments with optimization over raw pixels and GAN latent spaces in  Section \ref{ablations}.

\mypar{Implementation details.}
In \ours, we run the optimization loop for $N=160$ steps, which we found enough to transform most images. We use a resolution of 288 for encoding images with VQGAN, which compresses the images in a latent space with dimensions (256, 18, 18). 

We take advantage of  various pre-trained CLIP models, and combine their embeddings with concatenation, as shown in Figure~\ref{fig:clips}. 
By default, we use three image embedding networks with  different ResNet  and ViT architectures, which implement complementary inductive biases.
To encode an image with a single CLIP network, we average the embeddings of multiple augmentations of the input image (8 by default). We have empirically observed that using multiple augmentations per network stabilizes optimization in the early stages.

For the regularization coefficients, we use $\lambda_z=0.05,$ $\lambda_p = 0.15,\ \lambda_S=0.4,\ \lambda_I = 0.2$ as our default values. 
These coefficients are set using 
our ImageNet-based development set, and are fixed for all experiments. 

These implementation choices are analysed in Sec. \ref{hparam}.

\section{Experiments}

\begin{figure*}
    \centering
    \vspace{-2em}
    \includegraphics[width=.95\linewidth]{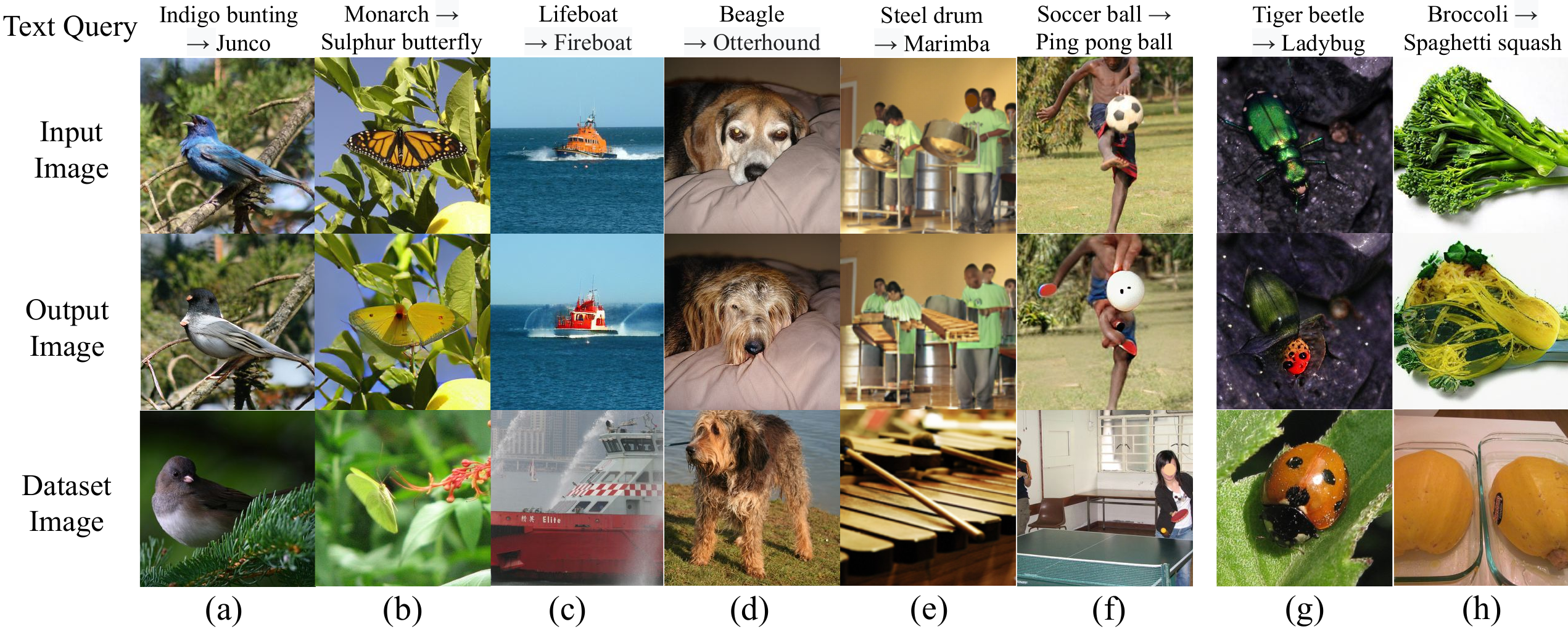}
    \caption{Transformation examples with \ours on ImageNet images. Columns (a)-(e) show examples of successful transformations. Column (f) shows an interesting behavior where another object has been added in the image to add more context (a table tennis racket in the hand of the person). The  last two columns  show the most frequent modes of failure: only part of the input object is transformed (g), or parts of the input object that should be changed are not changed: in column h, the transformed images still has a broccoli shape with green parts instead of an orange and round spaghetti squash).
    }
    \label{fig:visuresu}
    \vspace{-.5em}
\end{figure*}

Below, we first describe  our evaluation protocol in detail. 
We then present qualitative and quantitative results,  and an in-depth analysis of various components of our approach.

\subsection{Evaluation Protocol\label{eval}}

\mypar{Evaluation dataset.} 

We did not find a satisfying evaluation framework to study the problem of semantic image translation: existing dataset and metrics focus on narrow image domains, or random text transformation queries~\cite{li2020manigan,patashnik2021styleclip}. 
To overcome this, we have decided to build upon the ImageNet dataset~\cite{deng09cvpr} for its diversity and its high number of classes: by defining which class labels can be changed into one another (like \textit{cat} $\ra$ \textit{tiger}), we can build a set of sensible object-centric transformation queries. 
We have selected a subset of the 273 ImageNet labels that we manually split into 47 clusters according to their semantic similarity. For instance, there is a cluster containing all kinds of vegetables. Details on the subset selection and grouping are presented in the appendix.
We only consider  transformations  $S \rightarrow T$ where $S$ and $T$ are in the same cluster, in order to avoid nonsensical transformations between unrelated objects, \eg laptop $\ra$ butterfly.

For each target label $T$ we construct eight transformation queries by randomly sampling eight other classes $\{S_i\}$ within the same cluster, and sample a random image from each $S_i$ from the ImageNet validation set.
This gives  a total of 2,184 transformation queries that we split into a development set and a test set of equal size.
We use the development set to tune various hyper-parameters of our approach, and report evaluation metrics on the test set.

\mypar{Metrics.}

We  evaluate the success of the transformation by means of the \textbf{Accuracy} of an image   classifier, which is possible since we use ImageNet class labels as the transformation targets.

We use a DeiT~\cite{touvron20arxiv2} classifier, which has an ImageNet validation accuracy of 85.2\%. 
We judge a transformation  successful if, for the transformed image, class $T$ has the highest  probability among the  273 selected classes.

To assess naturalness of  transformed images, we use the Fréchet Inception distance (FID)~\cite{heusel2017gans}.
To avoid numerical instability related to estimating the feature distribution with a small number of samples, we use the ``Simplified FID'' ({\bf SFID}) \cite{kim2020simplified} which does not take into account the off-diagonal terms in the feature covariance matrix. In addition to the SFID, we use a class-conditional SFID score  ({\bf CSFID}) which is an average of the SFID scores computed for each target class separately.\footnote{Referred to as within-class FID in \cite{benny21ijcv}.} Because we compute these scores with a low number of examples for many classes, the CSFID score has a high bias, low variance profile on our dataset~\cite{chong2020effectively}, and we  
 have found it to be reliable and stable.
The CSFID metric is a measure of both image quality and transformation accuracy, as it measures the feature distribution distance between the transformed images and the reference images from the target class in the 
training set. 
Editing should not change parts of the image that are irrelevant to the transformation defined in the text,  \eg the background.  
We use the {\bf LPIPS} perceptual distance~\cite{zhang18cvpr}  to measure deviation from the input image.
It is a weighted $\ell_2$ distance of deep image features, and has been demonstrated to correlate well with human perceptual similarity. 
During training, we used the LPIPS distance using VGG features rather than AlexNet, so as to reduce bias in the evaluation results. 
The LPIPS  distance cannot differentiate between edits that are relevant to the text query, and those which are not; and we don't know the minimal LPIPS distance between an image and its closest successful transformation. Still, we argue that it should be as low as possible.

More details on the metrics are presented in appendix.

\subsection{Results}

\begin{figure}
    \centering
    \vspace{-1em}
    \includegraphics[width=\linewidth]{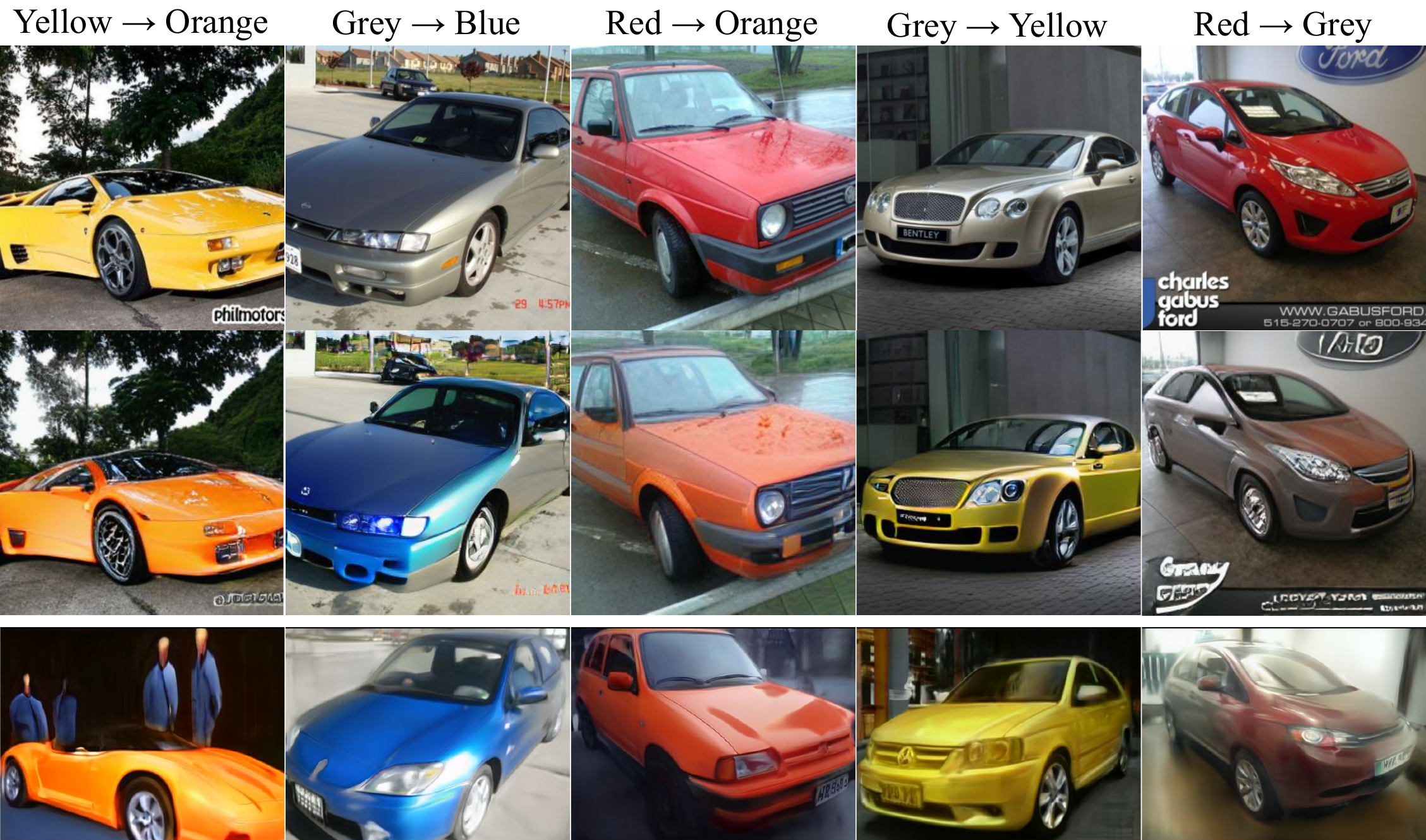}
    \caption{Example  transformations on the \textit{Cars} dataset:  input images (first row),  \ours results  (second row), StyleCLIP results based on a  StyleGAN2 backbone pre-trained on LSUN Cars.
    dataset (last row). Although GAN-based images have better details like the wheels, they are farther away from the input images.}
    
    \label{fig:cars30k}
\end{figure}

Qualitative results of FlexIT transformations on ImageNet images are presented in Figure \ref{fig:visuresu}, including successful transformations as well as several failure cases. 
To demonstrate the generality of our approach, we also  show examples of color transformations for images from the Stanford Cars dataset~\cite{KrauseStarkDengFei-Fei_3DRR2013} in Figure \ref{fig:cars30k}.

Semantic image translation is inherently a trade-off between having the most relevant and natural output image (as measured by Accuracy, CSFID and SFID), while staying as close as possible to the input image (as measured by LPIPS). 
We consider two extreme configurations as baselines,  which only optimize one of these two criteria: (i) The  \textsc{Copy} baseline, which simply copies the input image without any modification,
and  (ii) the \textsc{Retrieve} baseline that outputs a random validation image labelled with the target class $T$.
We add the \textsc{Encode} baseline that simply passes the input image through the VQGAN autoencoder.

We also evaluate  StyleCLIP~\cite{patashnik2021styleclip}, the most relevant text-driven image transformation algorithm from the literature. We consider the version most similar to our method that embeds images with an ImageNet-trained StyleGAN2,\footnote{We used the publicly available model from \url{https://github.com/justinpinkney/awesome-pretrained-stylegan2}, and  train our own e4e encoder~\cite{tov2021designing} to embed images into this latent space.} and iteratively updates the StyleGAN2 latent representation to maximize the similarity with a given text in the CLIP latent space. We have also trained ManiGAN~\cite{li2020manigan} on ImageNet with the official implementation.

\begin{table}
\centering
\small
\resizebox{\columnwidth}{!}{
\begin{tabular}{lrrrr}
\toprule
& \thead{LPIPS ↓} & \thead{Acc.\%↑} & \thead{CSFID ↓} & \thead{SFID ↓}  \\
\midrule
\textsc{Copy} & 0.0 & 0.45 & 106.0 & 0.20 \\
\textsc{Encode} & 17.5 & 1.6 & 107.5 & 2.99 \\
\textsc{Retrieve} & 72.4 & 90.6  & 27.2 & 0.23 \\
ManiGAN~\cite{li2020manigan} & 21.7 & 2.0 & 123.8 & 17.0 \\ 

StyleCLIP~\cite{patashnik2021styleclip}   & 33.4 & 8.0 & 146.6 & 35.8 \\ 
\ours (Ours) & 24.7 & 51.3 & 57.9 & 6.8 \\
\bottomrule
\end{tabular}}
\caption{Evaluation  of \ours and baselines on ImageNet images. 
}
\label{tab1}
\vspace{-2em}
\end{table}

Results are reported in Table~\ref{tab1}. 
As expected, the copy baseline is ideal on LPIPS and SFID, but fails to adapt to the transformation target $T$, and thus fails on Accuracy and CSFID.
The auto-encoding baseline fails on Accuracy and CSFID for the same reason, but demonstrates the non-trivial impact of using the VQGAN latent space on LPIPS and SFID. The \textsc{Retrieve} baseline provides ideal metrics for Accuracy, CSFID and SFID, as it returns natural images of the target class. 
It fails  on LPIPS, however, since the output image is unrelated to the input.

Our \ours approach  combines a  low LPIPS (24.7 \vs 17.5 for \textsc{Encode}) with an accuracy of 51.3\% and a CSFID of 57.9, which is closer to the CSFID of \textsc{Retrieve} (27.2) than that of \textsc{Encode} (107.5).
The StyleCLIP scores are poor, with high SFID and CSFID scores which was expected as StyleCLIP has been designed to work well where GANs shine.
The StyleGAN2 model we use, trained on ImageNet, is agnostic to class information and cannot synthesize realistic images for all ImageNet classes.
ManiGAN works well when trained on narrow domains with color change transformation requests, but we find that it does not produce convincing edits when trained on ImageNet.

\begin{figure}
    \centering
    \vspace{-1em}
    \includegraphics[width=\linewidth]{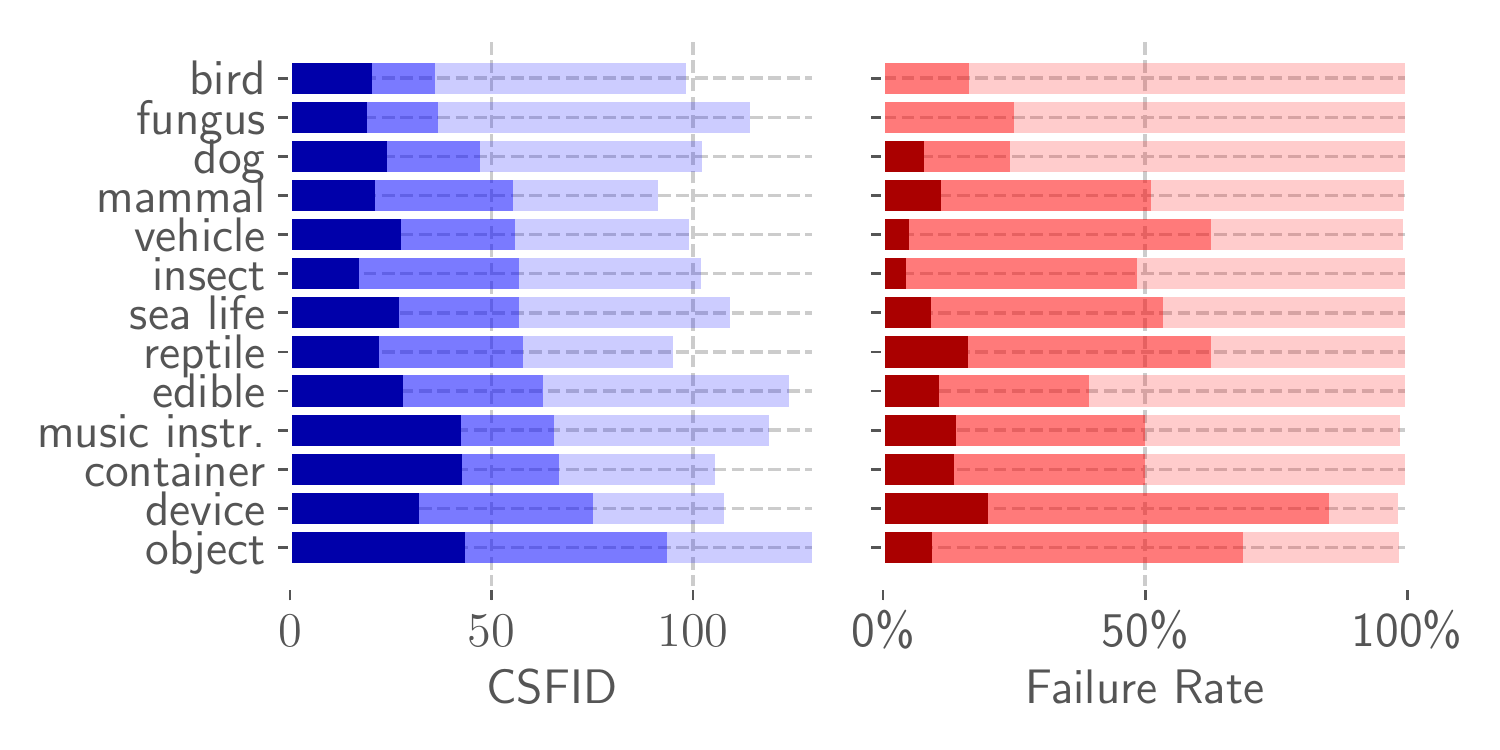}
    \caption{Groupwise CSFID  and Failure Rate (1-Accuracy),  lower is better for both metrics. 
    Dark colors: best possible values obtained with  \textsc{Retrieve} baseline; medium colors: scores obtained with \ours; light colors: values obtained with  \textsc{Copy} baseline. 
    }
    
    \label{fig:classwise}
\end{figure}

To provide  insight into which transformations work well, and which less so, we group the ImageNet clusters into 13 bigger groups (see appendix for details) and report the average CSFID and failure rate (1 - accuracy) scores for each group in Figure \ref{fig:classwise}.
Generally, transformations among  natural objects are more successful than transformations among man-made objects. We believe that this is mostly because the latter appear in a wider variety of shapes and contexts which leads to more difficult transformations.

\subsection{Ablation studies\label{ablations}}

\begin{figure}
    \centering
    \vspace{-1em}
    \includegraphics[width=\linewidth]{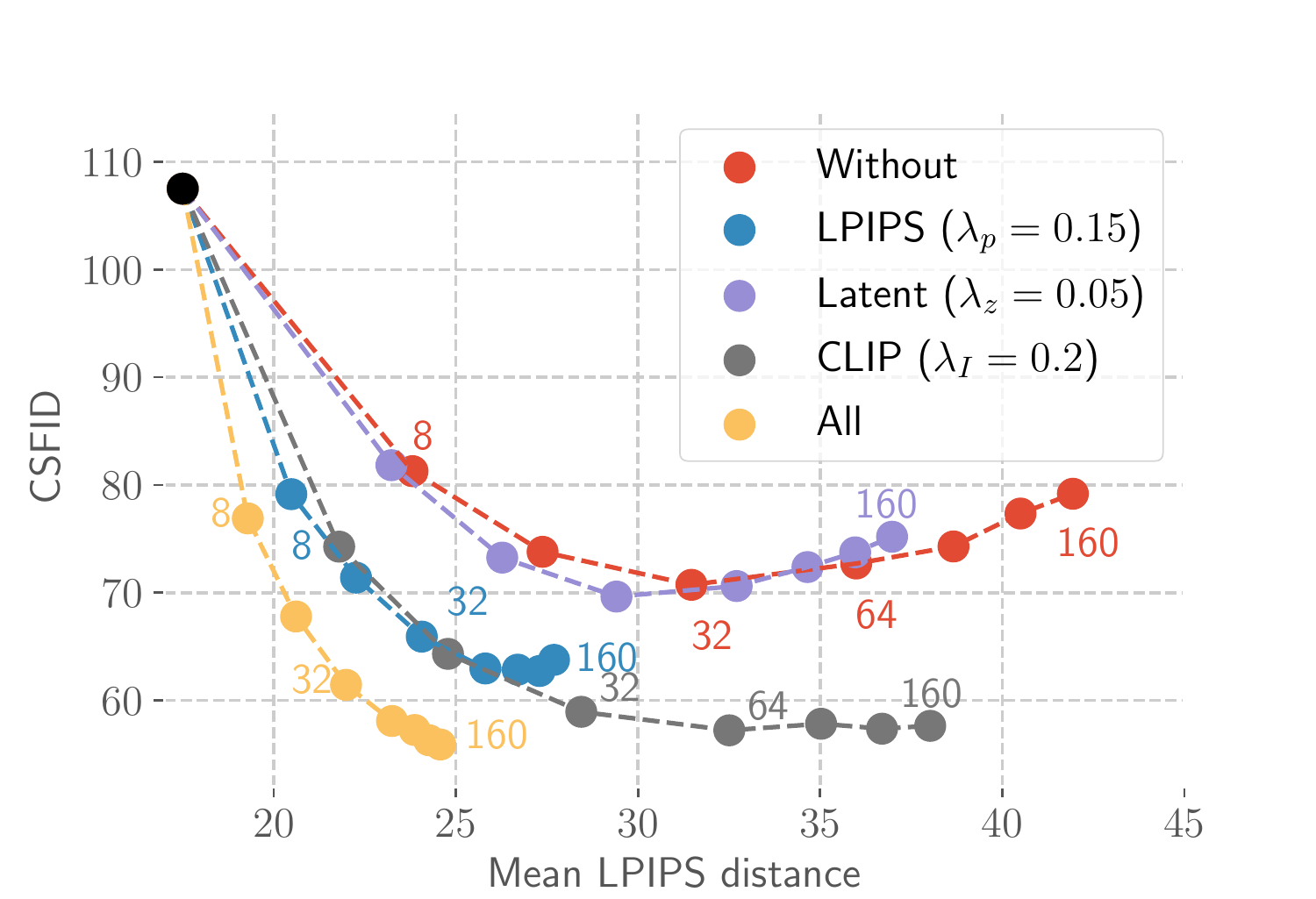}
    \caption{CSFID obtained without regularization, with individual LPIPS, Latent and CLIP regularizers, and using all. 
    Each curve corresponds to 160 steps of optimization on the dev.\ set. 
    }
    \label{fig:regul}
\end{figure}

\begin{figure}
    \centering
    \vspace{-1em}
    \includegraphics[width=\linewidth]{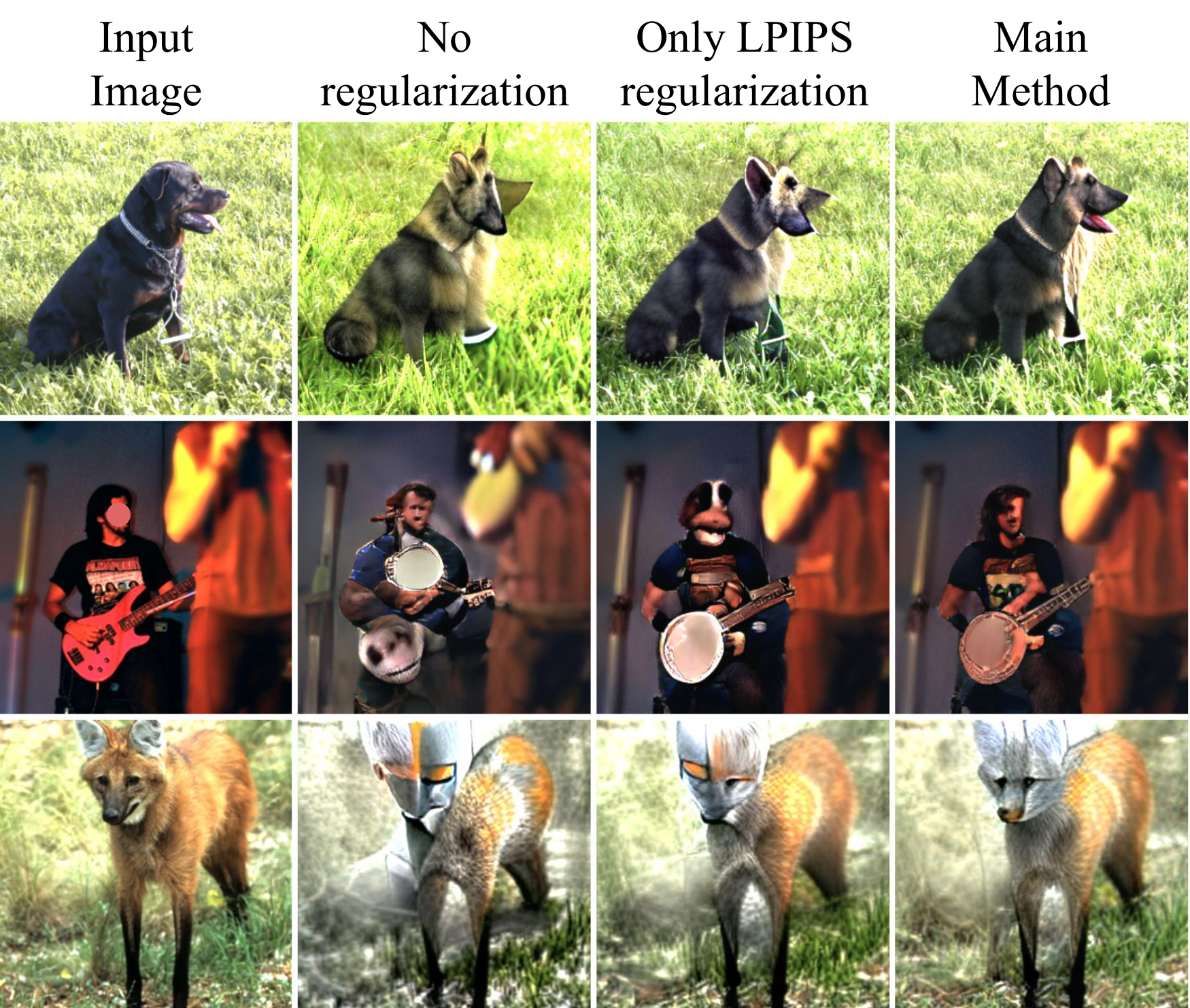}
    \caption{Example transformations with different regularizers. Textual queries from top to bottom: Rottweiler → German shepherd, Electric guitar → Banjo, Red wolf → Grey fox.
    }
    \label{fig:demo_reg}
\end{figure}

\mypar{Regularizers.}
In Figure \ref{fig:regul}, we show the evolution of CSFID along the optimization steps, where we consider our method without regularization, with each regularization scheme separately, and with all regularizers (default configuration).
Compared to not using regularization, the LPIPS regularization substantially improves the CSFID score along the optimization path, while also reducing LPIPS as expected. 
The CLIP regularizer has a similar effect, but is able to reduce  CSFID further while the LPIPS distance is only slightly reduced compared to our method without any regularization.
Using all regularizers allows us to obtain the lowest CSFID scores at low LPIPS. 
We believe that these two regularizers are complementary: while the LPIPS loss mitigates image deviation for local features, the CLIP loss provides semantic guidance which helps to reconstruct recognizable objects. 
Corresponding qualitative examples are shown in Figure \ref{fig:demo_reg}. 

\begin{figure}
    \centering
    \vspace{-1em}
    \includegraphics[width=.9\linewidth]{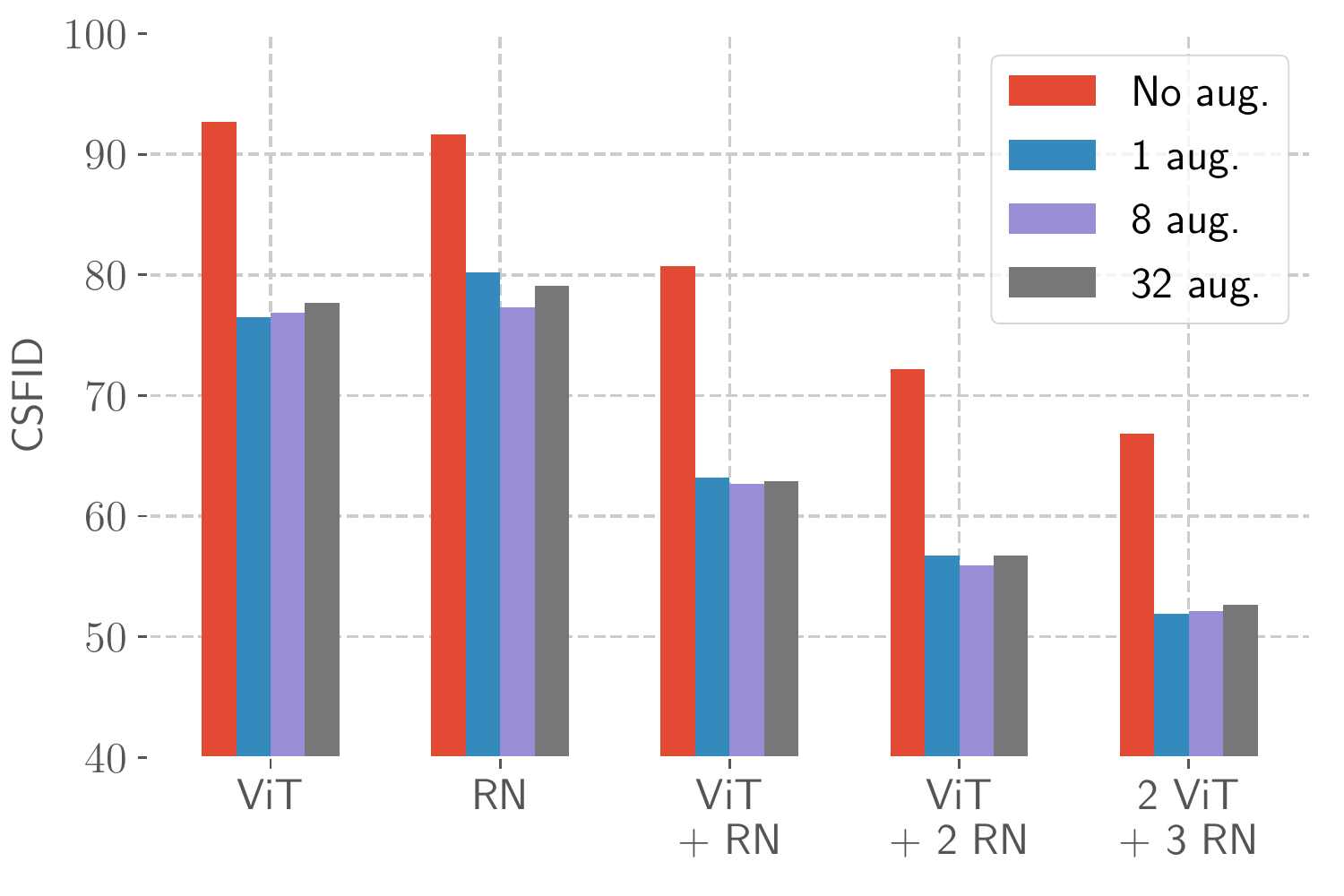}
    \caption{CSFID for different CLIP networks combinations and number of data augmentations options. Default setting: ViT+2RN.}
    \label{fig:augs}
    \vspace{-1em}
\end{figure}

\mypar{CLIP embedding module.} 
We study how different choices of CLIP image encoders impact the CSFID score. 
Our default configuration involves two ResNet-based networks and one ViT-based network to embed the image in the CLIP space. 
We experiment with a single ViT or ResNet, a combination of ViT with a single  ResNet,  
and also using all available pre-trained CLIP networks, which comprises a ViT-B/16, a ViT-B/32, a ResNet50, ResNet50x4 and ResNet50x16, see~\cite{radford21clip} for details on the modules.
For each CLIP network configuration, we experiment with  either  not using data augmentation, or using $d \in\{ 1, 8, 32\}$ augmentations.
We apply basic geometric augmentations that are commonly used to train image classification networks (more details in appendix).
Each of the $N_\textrm{nets}$ CLIP networks sees a different augmentation in each of the $N_\textrm{steps}$ steps of the optimization process,  resulting in a total of $d \times N_\textrm{nets} \times N_\textrm{steps}$ augmentations of the input image.

From the results in Figure \ref{fig:augs}, we see that while the  ViT and  ResNet embedding networks lead to similar results, they are complementary and combining them leads to a substantial improvement.  
Adding additional networks leads to further improvements.
Second, using data augmentation is very beneficial, and leads to a reduction in CSFID of 10 or more points for all network configurations. 
Using more than one augmentation does not improve results substantially: it suffices to a different augmentation for each network at each optimization step.
In our other experiments we use the three smallest (and fastest) CLIP networks as our default setting.

\begin{figure}
    \centering
    \vspace{-3em}
    \includegraphics[width=.9\linewidth]{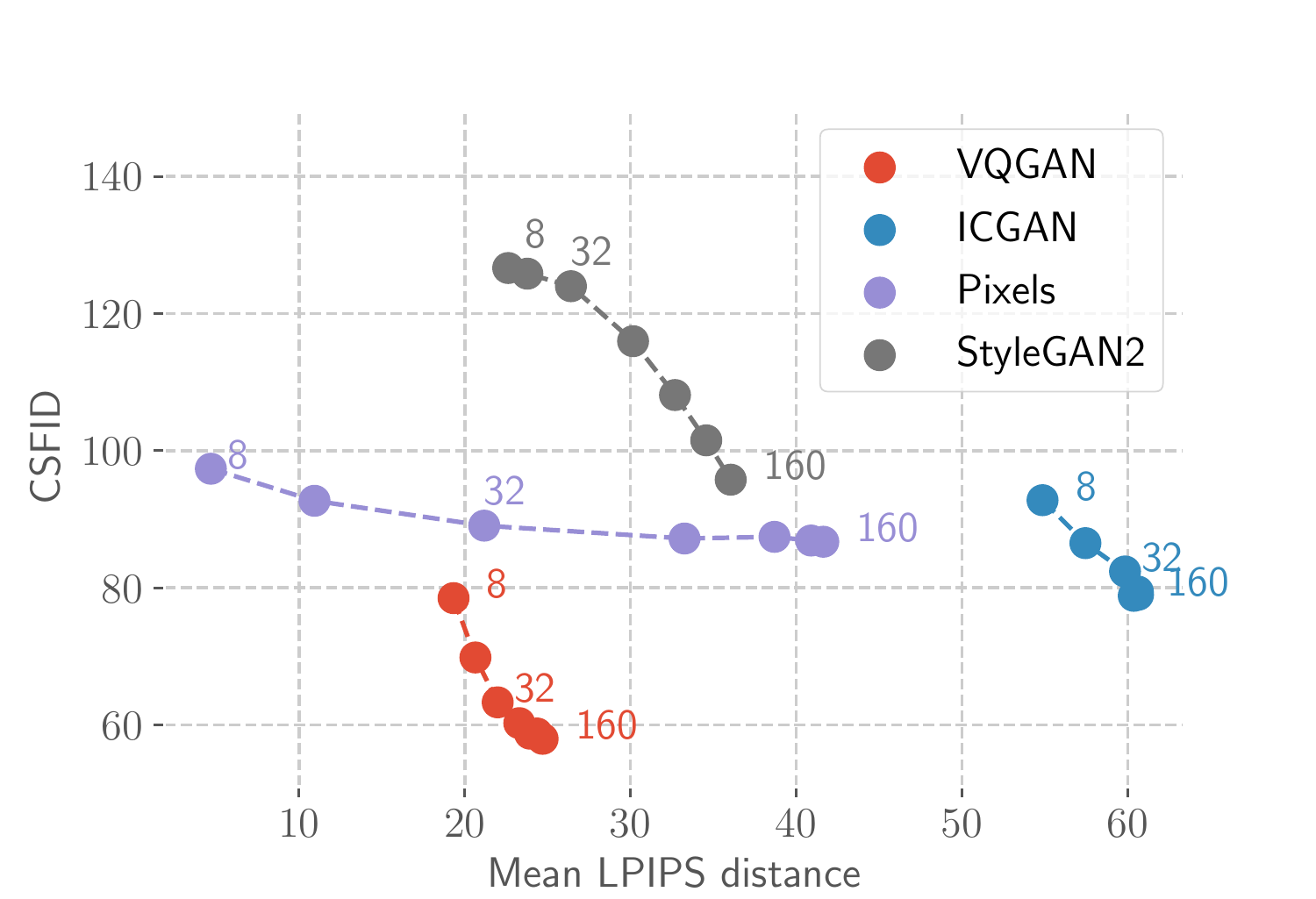}
    \caption{CSFID and LPIPS scores across iterations, using different latent spaces, or raw pixels, for optimization. 
    }
    \label{fig:encoders}
    \vspace{-1em}
\end{figure}

\mypar{Image optimization space.} 
We compare our choice of optimizing in the VQGAN latent space with using the latent spaces of StyleGAN2~\cite{karras20cvpr} and IC-GAN~\cite{casanova21nips}, as well as optimizing directly in the pixel space.
IC-GAN~\cite{casanova21nips}  generates images similar to an input image, 
and uses  a latent variable to allow for variability in its output.
As IC-GAN does not offer direct  inference of the latents  for a given image, we take 1,000 samples from the latent prior, and keep the one yielding minimal LPIPS distance to the input image. 
We found that  optimization to further reduce the LPIPS  \wrt the input image from this point on was not effective.
For StyleGAN2~\cite{karras20cvpr}, we use the same  network  pre-trained on ImageNet as we used for StyleCLIP.
To embed the evaluation images into this latent space, we first obtain an initial prediction of the vector with the e4e encoder~\cite{tov2021designing}, as in StyleCLIP, and then  perform an additional 1,000  optimization steps to better fit the input image, following the GAN inversion procedure described in \cite{karras19cvpr}.

The results in Figure \ref{fig:encoders} show that using the VQGAN latent space allows to substantially decrease the  CSFID score along the iterations, while only slightly increasing LPIPS. 
Using the raw pixel space is not effective to decrease the CSFID. 
IC-GAN has relatively good image synthesis abilities but it is hard to faithfully encode images in its latent space, yielding high LPIPS scores above 50. 
The StyleGAN2 latent space ($\mathcal{W+}$) is bigger, allowing generated images to be closer to the input images; however its CSFID scores are not competitive with the other approaches.

\subsection{Hyperparameter study\label{hparam}}

\begin{figure}
    \centering
    \vspace{-2em}
    \includegraphics[width=\linewidth]{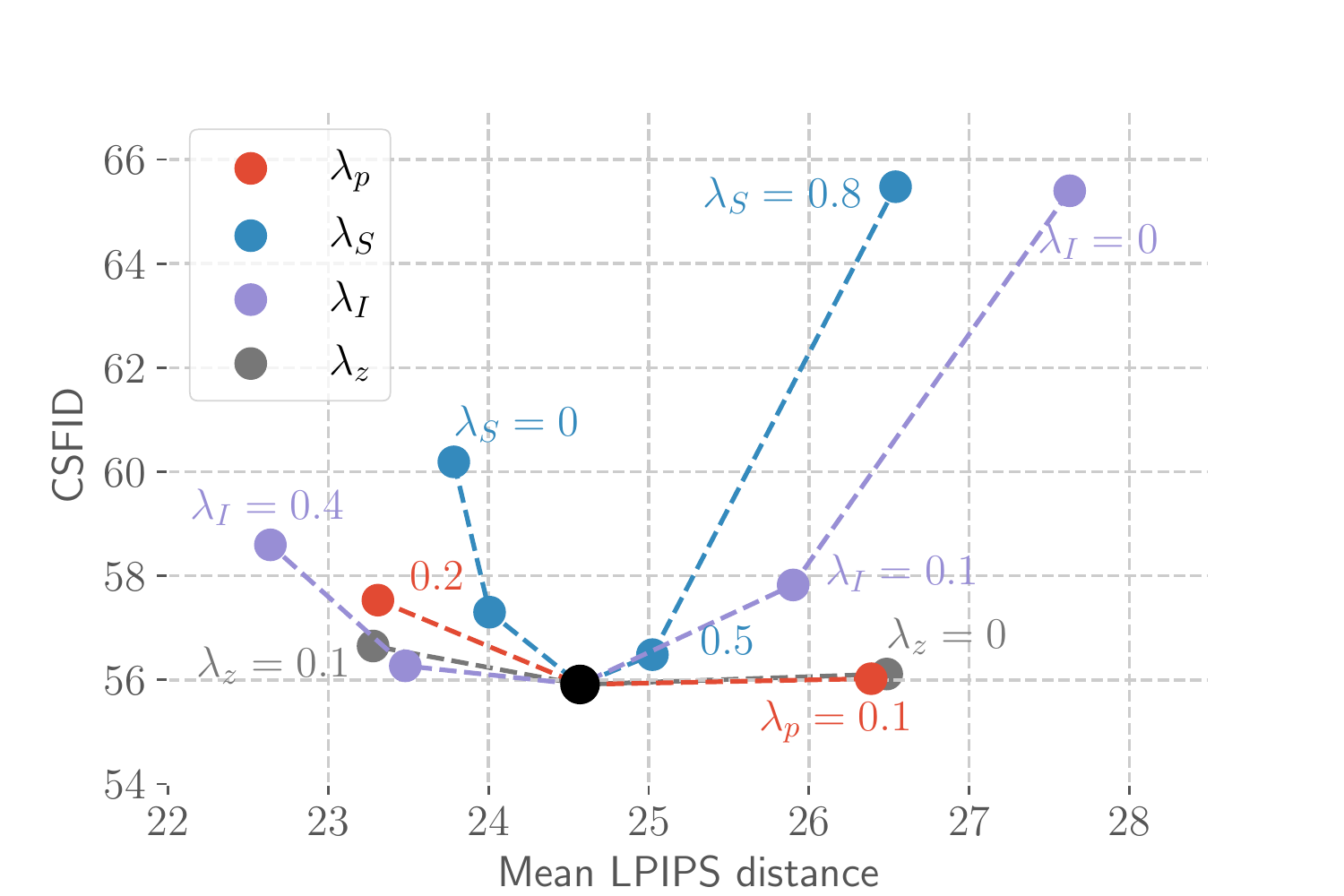}
    \includegraphics[width=\linewidth]{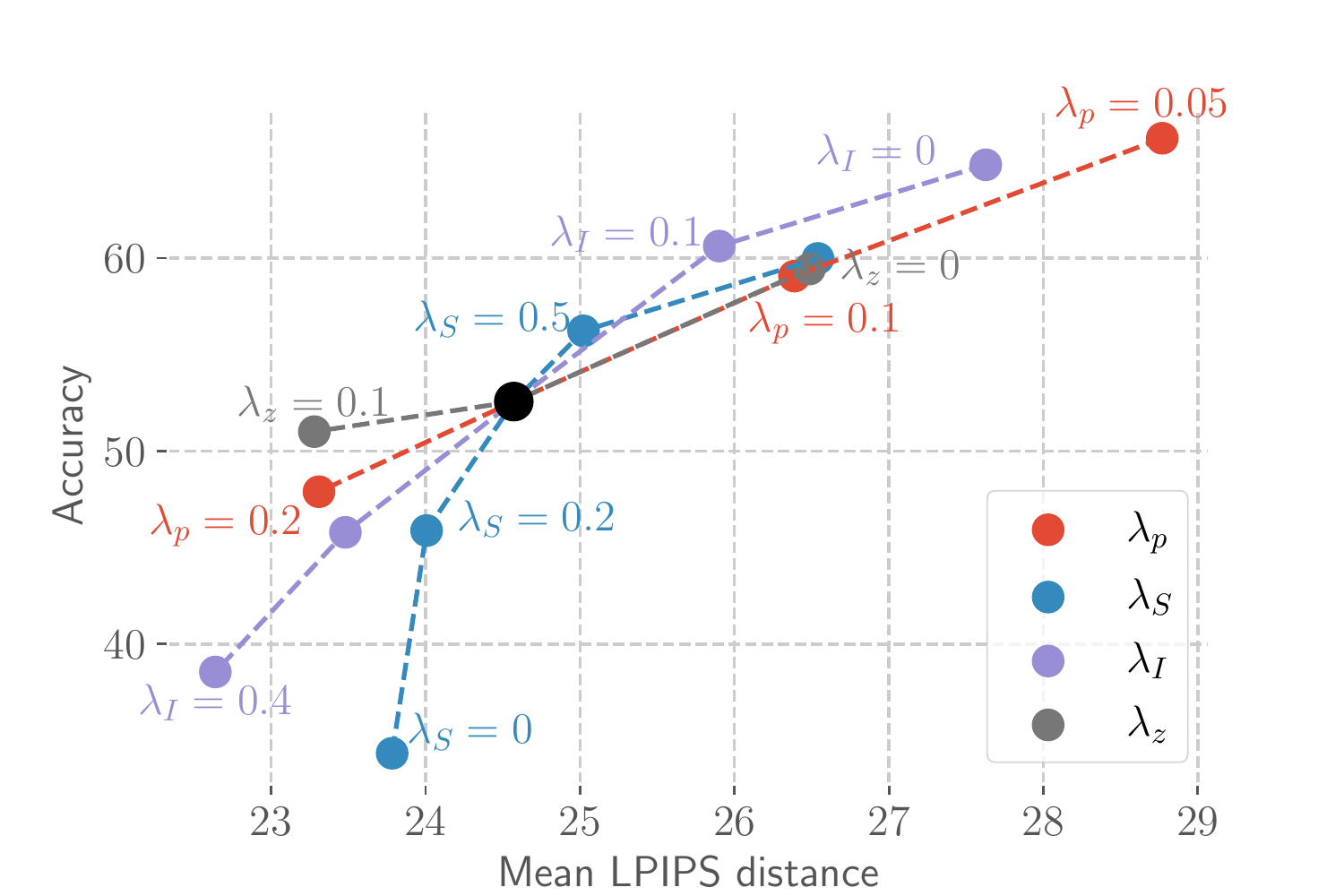}
    \caption{
    Effect  on CSFID and Accuracy of hyper-parameters;  default settings  represented by 
    the black dot, where all lines cross.
    }
    
    \label{fig:hparam_csfid}
    \vspace{-1em}
\end{figure}

In Figure \ref{fig:hparam_csfid}, we illustrate the effect of our hyper-parameters on the LPIPS, CSFID, and Accuracy metrics. 
For the three regularization parameters $\lambda_p, \lambda_z, \lambda_I$, we observe that
(i) the LPIPS distance with respect to the input image is smaller as the regularization gets stronger, as expected; 
(ii) less regularization allows more image modifications, yielding better accuracy scores, as illustrated in the bottom panel; 
(iii) there is a global minimum in CSFID scores when we make each hyperparameter vary independently (top panel). Regularization constraints are indeed useful to prevent inserting unnatural visual artifacts; however, too much regularization penalizes our algorithm as the distribution of output images gets closer to the input distribution, and thereby farther from the  target distribution.

The parameter $\lambda_S$, similarly to the regularization parameters, has a an optimal value which minimizes the CSFID. It is beneficial to give a hint to the optimization algorithm which semantic content should be changed, however focusing too much this objective reduces image realism.

For our main experiments, we set our hyper-parameters to  minimize the CSFID score on the development set. This is a natural choice given the convex shape of the CSFID scores, whereas optimizing for accuracy  would   remove the regularizers which is detrimental for image quality.

\section{Conclusion}

\mypar{Contributions.}
We propose \ours, a novel method for semantic image translation.
By relying on an autoencoder latent space, rather than specialized GAN latent spaces,
it can operate on 
a wide range of images. 
Using a general pre-trained multi-modal embedding space provides  flexibility, giving \ours the ability to process free-text transformation queries without training. We  also propose an evaluation protocol for semantic image translation, based on ImageNet, which we use to thoroughly evaluate our approach and its components.

\mypar{Limitations.}
Our method works best for semantic translation when the input image provides guidance, but has difficulties synthesizing realistic novel objects from scratch. 
Also, while we studied transformations that change the class or color of the main object in a scene, other transformations of interest could consider changing the action of a subject (person walking \vs  running), changing object attributes, adding or deleting objects, or consider more elaborate textual descriptions which require non-trivial grounding in the image (``change the color of car parked next to the bicycle.''). However, progress in this direction will require to identify the right data and evaluation metrics. 

\mypar{Broader impacts.}
As our algorithm relies on CLIP for editing, it could potentially inherit biases embedded in the CLIP model. The authors of CLIP have demonstrated that similarly to other neural network models, CLIP is subject to fairness issues such misclassifying human faces into non-human or crime-related categories, and producing gender biased associations where some labels that describe high status occupations are disproportionately more attached to images of men than that of women. Our editing method could reflect such biases if prompted transformations such as doctor $\rightarrow$ newscaster, although we have not observed experimental evidence of this. A potential bias mitigation strategy would be to add constraints with CLIP prompts, for instance by enforcing that the probability of the labels \textit{man} and \textit{woman} remain the same before and after editing.

Our model provides new capabilities to an expanding set of image editing and synthesis tools based on deep generative models. As any generative image model, synthetic images generated by our method can potentially be used in unintended ways with undesirable effects. We believe however that open publication of research in this area contributes to a good understanding of such techniques, and can aid the community in efforts to develop method that detect unauthentic content.


{\small
\bibliographystyle{ieee_fullname}
\bibliography{mybib, gco, jjv}
}
\clearpage

\appendix

\section{Appendix}

\subsection{Assets}
We provide a list of the assets used in our work (datasets, code, and models) in Table \ref{tab:links} (links) and Table \ref{tab:licences} (licences).

\begin{table*}
\center
\begin{tabular}{lll}
\toprule
\bf Asset Name & \bf Link \\
\midrule
ImageNet & https://www.image-net.org \\
Cars & https://ai.stanford.edu/~jkrause/cars/car\_dataset.html \\
LPIPS & https://github.com/richzhang/PerceptualSimilarity \\
FID & https://github.com/mseitzer/pytorch-fid \\
DeiT & https://github.com/facebookresearch/deit \\
CLIP & https://github.com/openai/CLIP \\
VQGAN & https://github.com/CompVis/taming-transformers \\
IC-GAN & https://github.com/facebookresearch/ic\_gan \\
StyleGAN2 & https://github.com/justinpinkney/awesome-pretrained-stylegan2 \\
e4e & https://github.com/omertov/encoder4editing \\
\bottomrule
\end{tabular}
\caption{List of asset links.}
\label{tab:links}
\end{table*}

\begin{table*}
\center
\begin{tabular}{lll}
\toprule
\bf Asset Name & \bf Asset type & \bf License \\
\midrule
ImageNet & Images & https://www.image-net.org/download.php \\
Cars & Images & https://ai.stanford.edu/~jkrause/cars/car\_dataset.html\\
LPIPS & Code and Models & BSD-2-Clause License \\
FID & Code and Models & Apache-2.0 License \\
DeiT & Code and Models & Apache License 2.0 \\
CLIP & Code and Models & MIT License \\
VQGAN & Code and Models & MIT License \\
IC-GAN & Code and Models & Attribution-NonCommercial 4.0 International \\
StyleGAN2 & Code and Models & https://github.com/justinpinkney/awesome-pretrained-stylegan2 \\
e4e & Code &  MIT License \\
\bottomrule
\end{tabular}
\caption{List of asset licenses.}
\label{tab:licences}
\end{table*}

\subsection{Datasets}
\label{appendix.datasets}

To design transformation queries from ImageNet classes, we have grouped classes into clusters by semantic similarity, upon manual inspection of the WordNet hierarchy of classes. 
The resulting clusters are shown in Table \ref{fig:clusters}. This process resulted in 273 classes gathered in 47 clusters. We have not included all ImageNet classes because (i) we wanted to reduce the large number of dog breed classes, and (ii) a lot of classes were ``standalone classes" with no natural target for transformation among the other classes. The clusters are then grouped into 13 bigger ``groups'' that are used solely for visualization in Figure 6 
 of the main paper.

\subsection{Evaluation metrics}

\mypar{LPIPS.} 
As recommanded by the authors of~\cite{zhang18cvpr}, we use the   AlexNet~\cite{krizhevsky12nips} backbone to compute LPIPS distance, when we use it as an evaluation metric. 
To avoid using the same metric for optimization,  we compute the LPIPS in the perceptual regularization term $\mathcal{L}_{perc}$, see Eq.\ (3) of the main paper,  using the VGG16 network~\cite{krizhevsky12nips}.

The LPIPS distance is computed at an image resolution of 256, for both evaluation and optimization.
In the main paper, all LPIPS scores have been multiplied by 100 for readibility.

\mypar{(C)SFID.}
The FID metric~\cite{heusel17nips} measures the distance between the distributions of the real images and generated images in the feature space of an InceptionV3 classifier \cite{szegedy16cvpr}.

More formally, let $\mu^r$ and $\sigma^r$ be the mean and standard deviation of inception features for the real images, and $\mu^s$ and $\sigma^s$ for the synthetic images. The Simplified FID \cite{kim2020simplified} is computed as 

\begin{equation}
SFID(\alpha) = \lVert \mu^r - \mu^s \rVert ^2 + \alpha \lVert\sigma^r - \sigma^s\rVert^2.
\end{equation}

It does not take into account the off-diagonal terms in the feature covariance matrix to avoid numerical instability.

The  Conditional Simplified FID (CSFID) is computed in the same manner but for each target class separately, and then averaging the resulting scores: With $\mu^r_c$ and $\sigma^r_c$ the mean and standard deviation of inception features for the real images belonging to class $c$, and $\mu^s_c$ and $\sigma^s_c$ for the synthetic images, we have
\begin{equation}
CSFID(\alpha) = \frac{1}{|C|} \sum_c \lVert \mu^r_c - \mu^s_c \rVert ^2 + \frac{\alpha}{|C|} \sum_c \lVert\sigma^r_c - \sigma^s_c\rVert^2.
\label{eq:sfid}
\end{equation}

 We have noticed that the distance on standard deviations was not very discriminative: since we are modifying images and not generating images from scratch, we already have a lot of diversity in the generated images. Experimentally, using $\alpha > 0$ mostly consisted in adding a bias term in this metric, therefore we chose to use $\alpha =0 $ in the (C)SFID scores. 

Since the images we transform are extracted from the ImageNet validation set, we use the ImageNet training set as our reference distribution to compute the (C)SFID scores. 
As for LPIPS, the (C)SFID scores are computed at an image resolution of 256.

\mypar{Accuracy.} We use a DeiT classifier \cite{touvron20arxiv2} trained on ImageNet, which takes images of size 384$\times$384. Smaller images are upsampled before being passed to the classifier.

\subsection{Details on the multimodal encoder}

For data augmentations, we use a random horizontal flipping and a random rotation between $-10$ and $10$ degrees, followed by cropping the image (keeping at least $80\%$ of the input image) with aspect ratio between $0.9$ and $1.1$.
For the CLIP-based multimodal encoders, we have considered all CLIP networks freely available,  listed in Table \ref{tab:clip}.

\begin{table}[H]
\center
\begin{tabular}{lcc}
\toprule
\bf Backbone & \bf Params. & \bf Latent dim.\\
\midrule
RN50 & 38M & 512\\
RN50x4 & 87M & 640\\
ViT-B/32 & 88M & 512\\
ViT-B/16 & 86M & 512 \\
RN50x16 & 167M & 768\\
\bottomrule
\end{tabular}
\caption{Visual backbones used for the multimodal encoder. Our default configuration only includes the ViT-B/32, the RN50 and the RN50x4.}
\label{tab:clip}
\end{table}

\subsection{Additional qualitative results}

In Figure \ref{fig:encoders}, we show qualitative results when we replace the VQGAN image encoder with other GAN-based encoders. VQGAN has a native encoder and decoder, and thus the initial latent vector is obtained directly. For StyleGAN2 \cite{karras20cvpr}, we use the e4e encoder \cite{tov2021designing} followed by an additional 1,000 steps of LPIPS minimization. For the IC-GAN \cite{casanova21nips} model, we use the BigGAN \cite{brock19iclr} backbone as generator. IC-GAN is naturally conditioned on the SwaV embedding \cite{caron20nips} of the input image; for added robustness we sample 1,000 latent points and choose the one yielding smallest LPIPS distance with respect to the input image. 

Figure~\ref{fig:steps} shows intermediate transformation results with FlexIT for 0, 8, 16, 32 and 160 optimization steps. 
The result after zero optimization steps shows the effect of autoencoding the input image, without changing the latent representation.
Figure \ref{fig:failures} show representative failure cases for our method, due to either the regularization method or the multimodal embedding space. Finally, in Figure~\ref{fig:cars} we show  additional color transformation results on the Cars30k dataset.

\begin{figure*}[t!]
    \centering
    \includegraphics[width=0.9\linewidth]{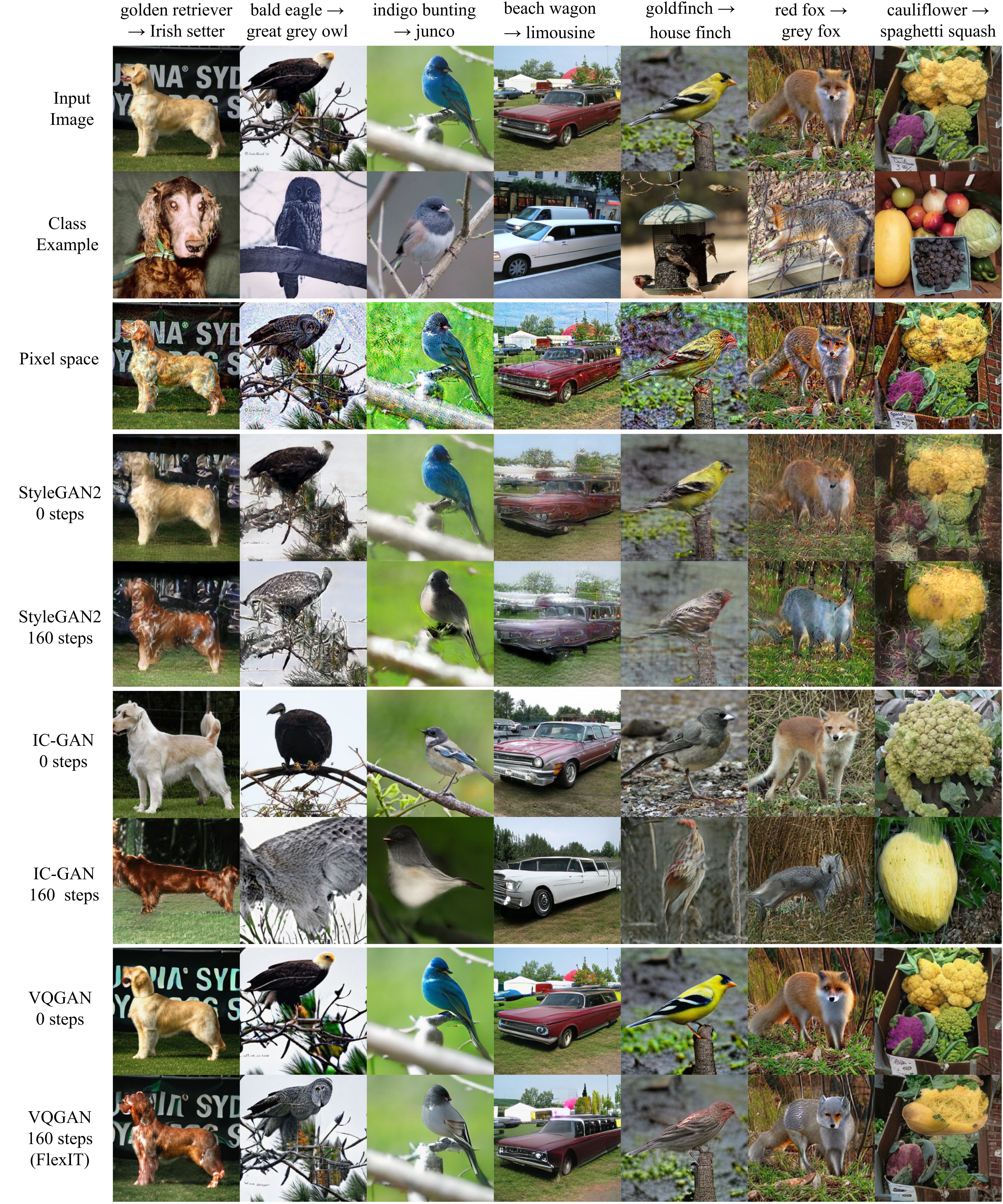}
    \caption{
    Transformation examples with various backbones for the image latent space. For each latent space, we show the initial image decoded from the initial point $z_0$, and the resulting image after 160 optimization steps. 
    The three latent spaces differ substantially in their encoding images (0 steps). The IC-GAN latent space provides natural images that are far away from the input image due to the limited generator capacity in conjunction with the smaller latent space size (2560 dim.). StyleGAN2 images preserve the input image appearance thanks to the larger size of its latent space $\mathcal{W+}$ (8192), however images contain many unnatural artifacts due to the challenges of embedding images in this latent space \cite{tov2021designing}. The VQGAN latent space leads to the best  reconstruction results.
    After 160 steps of optimization, the images generated with StyleGAN2 still have the same unnatural artifacts, and images generated with IC-GAN remain natural but  far from the input images.
    VQGAN, which we use in FlexIT, achieves good edits while preserving the overall image appearance.
    The pixel-space method introduces high-frequency artifacts, without substantially modifying the high-level semantic image content, resembling adversarial examples for image classification. 
    }
    \label{fig:encoders}
\end{figure*}

\begin{figure*}[t!]
    \center
    \includegraphics[width=\linewidth]{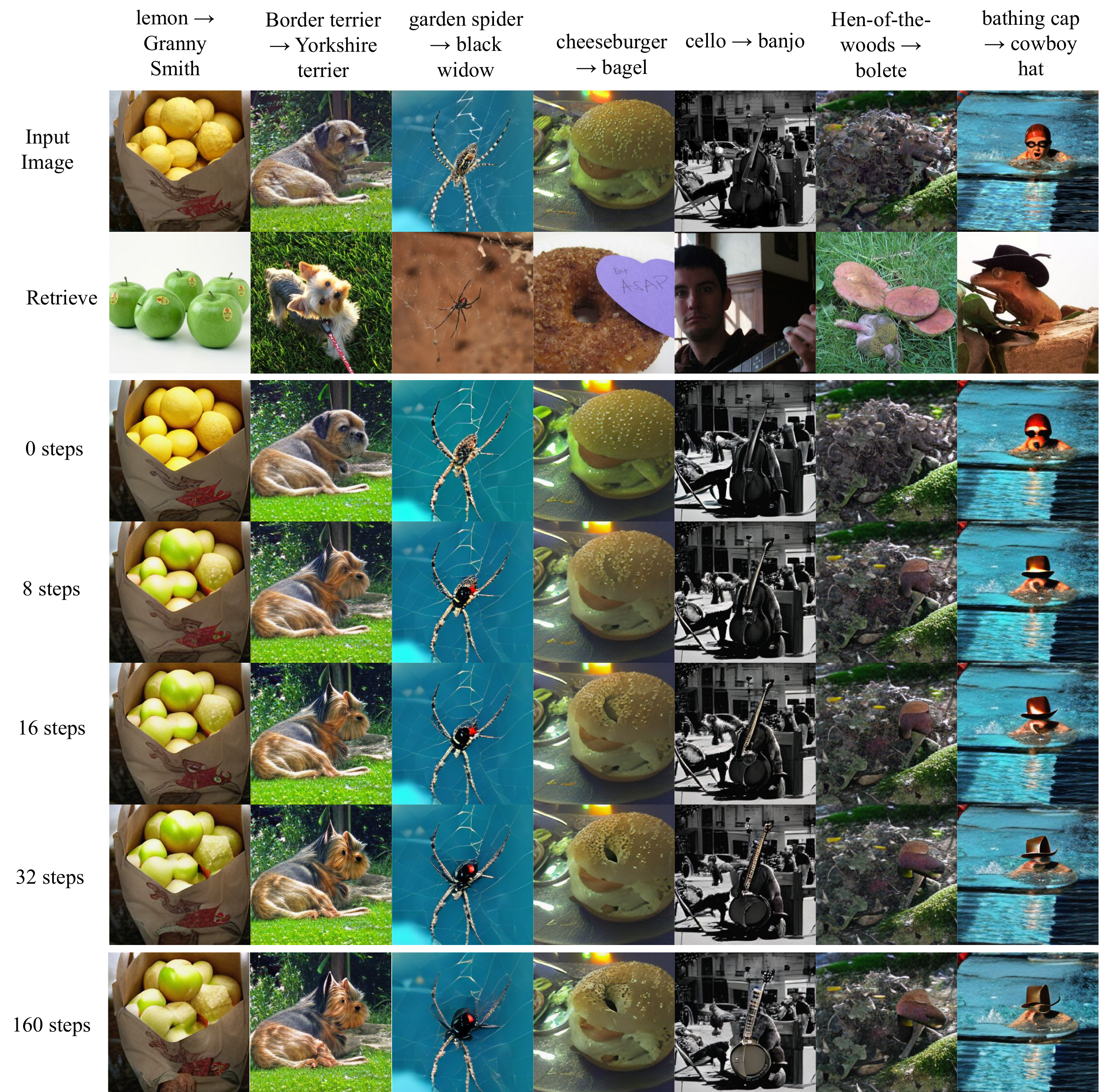}
    \caption{Intermediate transformation results obtained with FlexIT. Note that most edits only require 32 steps to be completed; some edits benefit from longer optimization schemes, such as the  spider and the banjo.
    }
    \label{fig:steps}
\end{figure*}

\begin{figure*}[t!]
    \center
    \includegraphics[width=\linewidth]{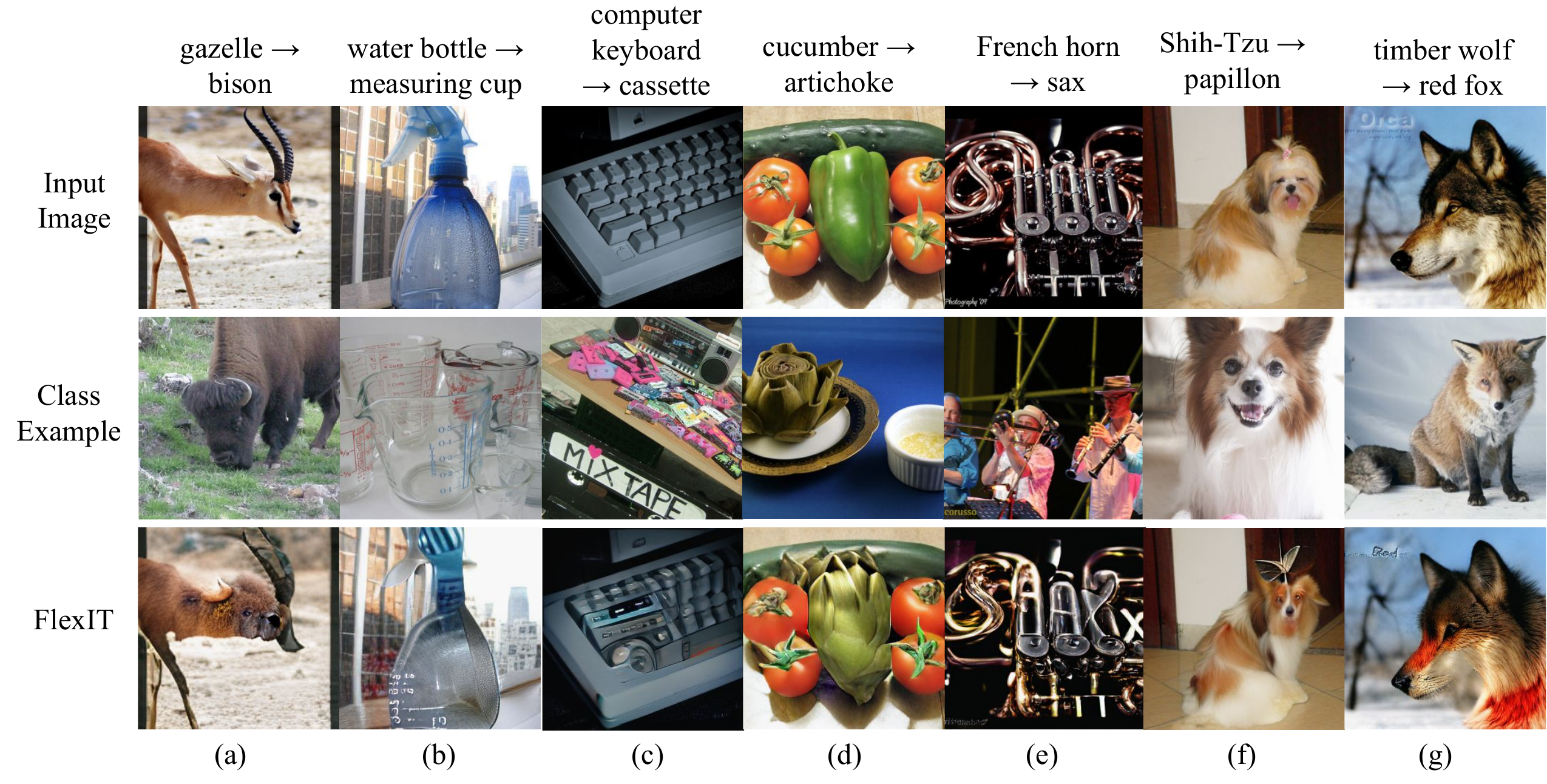}
    \caption{
    Representative failure cases of FlexIT. 
    The first three columns show examples where the regularization with respect to the initial image was too strong. (a): FlexIT added bison-like  texture but fails to change  the shape convincingly.
    (b): markings have been added to the  bottle, but without changing its  shape to that  of a measuring cup. 
    (c): only a part of the input object was changed. 
    (d): the bell pepper rather than the cucumber was transformed, probably because the former is more centered, and has a better initial shape. 
    Columns (e)--(g) show failure cases related to the CLIP embedding space. 
    (e): we observe an interesting text synthesis behaviour where the letters of the target class ``sax'' have been written in the image. This is related to the OCR capabilities of CLIP. 
    (f): a butterfly is synthesized on the head of the dog (CLIP optimized for both the dog breed papillon and the insect papillon). 
    (g): an unrealistic image is produced by adding saturated red to the image.
    }
    \label{fig:failures}
\end{figure*}

\subsection{Ablation results}
In Table \ref{table:ablations}, we show ablation experiments for all FlexIT parameters; this includes the CSFID scores of the hyper-parameter configurations reported in Figure 11 of the main paper.

In Table \ref{table:runtime}, we show ablations for combining multiple CLIP networks and using multiple data augmentations in the multimodal encoder. This includes the CSFID scores reported in Figure 9 of the main paper; we also report the runtime needed for each algorithm.

\begin{table}
\small
\center
\begin{tabular}{lrrrr}
\toprule
 & \bf Acc.↑ & \bf  LPIPS↓ & \bf  CSFID↓ & \bf  SFID↓ \\
\midrule
                   $\lambda_I=0$ & \bf 64.8 &   27.6 &   65.4 & 12.3 \\
                   $\lambda_I=0.1$ & 60.6 &   25.9 &   57.8 &  8.3 \\
                     
                    \rowcolor{LightGrey} $\lambda_I=0.2$   & 52.6 &   24.6 &   \bf 55.9 &  6.4 \\
                   $\lambda_I=0.3$ & 45.8 &   23.5 &   56.3 &  5.5 \\
                   $\lambda_I=0.4$ & 38.6 &   \bf 22.6 &   58.6 &  \bf 5.0 \\
                   \midrule
                   $\lambda_S=0.0$ & 34.3 &   \bf 23.8 &   60.2 &  \bf 4.8 \\
                  $\lambda_S=0.2$ & 45.9 &   24.0 &   57.3 &  5.5 \\
                  \rowcolor{LightGrey} $\lambda_S=0.4$   & 52.6 &   24.6 &   \bf 55.9 &  6.4 \\
                  
                   $\lambda_S=0.5$ & 56.2 &   25.0 &   56.5 &  7.1 \\
                   $\lambda_S=0.8$ & \bf 60.0 &   26.5 &   65.5 & 11.7 \\
                   \midrule
                   $\lambda_z=0.0$ & \bf 59.4 &   26.5 &   56.1 &  7.1 \\
                   \rowcolor{LightGrey} $\lambda_z=0.05$& 52.6 &   24.6 &  \bf 55.9 &  6.4 \\
                   $\lambda_z=0.1$ & 51.0 &   \bf 23.3 &   56.7 &  \bf 6.3 \\
                   \midrule
                   $\lambda_p=0.05$ & \bf 66.2 &  28.8 &  \bf 56.0 &  7.9 \\
                   $\lambda_p=0.1$ & 59.1 &   26.4 &   \bf 56.0 &  7.2 \\
                   \rowcolor{LightGrey} $\lambda_p=0.15$   & 52.6 &   24.6 &  \bf 55.9 &  6.4 \\
                   $\lambda_p=0.2$ & 47.9 &   \bf 23.3 &   57.5 & \bf  6.3 \\
                   \midrule
                    $\ell_1$ & \bf 54.2 &   \bf 24.6 &   56.3 &  6.5 \\
                    $\ell_2$ & 52.4 &   \bf 24.5 &  \bf 55.9 &  6.8 \\
                    \rowcolor{LightGrey} $\ell_{2,1}$ & 52.6 &   \bf 24.6 &  \bf  55.9 &  \bf 6.4 \\
                    \midrule
                    $lr=0.025$ & 47.6 &   \bf 22.5 &   58.3 &  \bf 6.0 \\
                    \rowcolor{LightGrey} $lr=0.5$     & 52.6 &   24.6 &    55.9 &  6.4 \\
                    $lr=0.1$ & \bf 60.4 &   27.6 &   \bf 54.8 &  7.2 \\
                    \midrule
                  resolution 256 & 53.8 &   24.8 &   56.8 &  7.2 \\
                  \rowcolor{LightGrey} resolution 288 & \bf 52.6 &   24.6 &   \bf 55.9 &  \bf 6.4 \\
                  resolution 320 & 54.3 &   \bf 24.0 &   57.4 &  7.3 \\
                  \bottomrule
\end{tabular}

\caption{\label{table:ablations} FlexIT ablation results. $lr$ is the learning rate. Lines corresponding to our default configuration are marked in light grey. The norms $\ell_1$,  $\ell_2$, and $\ell_{2,1}$  refer to the distance used for regularization in the VQGAN latent space. Best values for each metric are shown in bold inside each group of parameter values.
}
\end{table}

\begin{table}
\small
\center
\begin{tabular}{lrrrrrr}
\toprule
 networks & d & \bf Acc.↑ &  \bf LPIPS↓ &  \bf CSFID↓ &  \bf SFID↓ & \bf \thead{sec.\\ /im} \\
 
\midrule
         ViT-B/32 & 0 & 9.4 &   21.8 &   92.7 &  7.4 & 27s   \\
         ViT-B/32 & 1 & 37.5 &   26.4 &   76.5 & 11.1 & 27s  \\
         ViT-B/32& 8 & 35.1 &   25.4 &   76.9 & 10.7 & 33s \\
         ViT-B/32& 32 & 35.5 &   25.0 &   77.7 & 10.8 & 53s \\
\midrule
        RN50x4 & 0 & 13.4 &   23.8 &   91.6 & 11.8 & 35s \\
         RN50x4& 1 & 32.5 &   27.4 &   80.2 & 13.7 & 35s  \\
         RN50x4& 8 & 31.0 &   25.2 &   77.3 & 12.3 & 53s \\
        RN50x4& 32 & 27.0 &   24.2 &   79.1 & 11.7 & 122s \\
\midrule

          2 nets & 0 & 23.0 &   22.8 &   80.7 &  9.5 & 39s \\
          2 nets & 1 & 50.6 &   26.4 &   63.2 &  8.9 & 39s \\
          2 nets & 8 & 47.8 &   24.9 &   62.7 &  8.4 & 64s \\
          2 nets & 32 & 47.4 &   24.2 &   62.9 &  8.1 & 160s \\
  \midrule
        3 nets & 0 & 30.4 &   22.5 &   72.2 &  8.3  & 45s\\
        3 nets& 1 & 54.9 &   26.0 &   56.7 &  6.7 & 45s \\
        \rowcolor{LightGrey}
        3 nets& 8 & 52.6 &   24.6 &   55.9 &  6.4 & 75s \\
        3 nets& 32 & 51.7 &   24.0 &   56.7 &  6.7 & 190s \\
        \midrule
         5 nets & 0 & 39.6 &   22.4 &   66.8 &  7.7 & 70s \\
        5 nets& 1 & 60.3 &   25.5 &   51.9 &  5.5 & 70s \\
        5 nets& 8 & 60.1 &   23.9 &   52.1 &  5.4 & 176s \\
        5 nets& 32 & 52.0 &   22.8 &   52.7 &  5.2 & 560s \\
        \bottomrule

\end{tabular}

\caption{\label{table:runtime} Ablation results for the multimodal encoder components. $d$ is the number of augmentations. $d=0$ means that the encoder takes the unchanged image as input; For $d=1$, the encoder takes only one (augmented image), which explains why the edit time is the same as $d=0$.
When considering $n$ CLIP networks, we take the first $n$ elements in the following list: RN50x4, ViT-B/32, RN50, ViT-B/16, RN50x16. 
Our default configuration  is marked in light grey. 
Last column gives computation time per image in seconds.
}

\end{table}

\subsection{Results on ManiGAN evaluation setup}

Evaluating text-driven image editing requires (1) a list of sensible transformation queries, and (2) a method for evaluating the quality and accuracy of the generated result. The evaluation protocol in ManiGAN consists in (1) choosing \textit{random} COCO captions/image pairs and thus leading to noisy transformations and (2) calculating the image-text similarity score which was used as a loss term during their training, leading to bias in the final scores. In the main paper, we compare different methods using our novel evaluation protocol, which was carefully designed to avoid these pitfalls. Nonetheless, we show in Tab.\ \ref{results_manigan} that even with the ManiGAN protocol, FlexIT improves upon the ManiGAN scores by a large margin.

\begin{table}[H]
\small
\centering
\begin{tabular}{l|l|l|l|l|}
\cline{2-5}
                              & \multicolumn{1}{c|}{IS↑} & \multicolumn{1}{c|}{SIM↑} & \multicolumn{1}{c|}{DIFF↓}             & \multicolumn{1}{c|}{MP↑} \\ \hline
\multicolumn{1}{|l|}{ManiGAN} & 14.96                   & 0.087                    & 0.216                                 & 0.068                   \\ \hline
\multicolumn{1}{|l|}{FlexIT}  & \textbf{18.19}          & \textbf{0.177}           & {\color[HTML]{000000} \textbf{0.146}} & \textbf{0.151}          \\ \hline
\end{tabular}
\caption{ManiGAN evaluation on random edits from COCO.}
\label{results_manigan}
\end{table}

\renewcommand\cellgape{\Gape[1pt]}
\begin{table*}
\small
\center
\begin{tabular}{lll}
\toprule
\thead[l]{Group} & \thead[l]{Cluster} & \thead[l]{Classes} \\
\midrule
bird & bird of prey & bald eagle, kite, great grey owl \\
bird & finch & indigo bunting, goldfinch, house finch, junco \\
bird & grouse & black grouse, prairie chicken, ptarmigan, ruffed grouse \\
bird & seabird & king penguin, albatross, pelican, European gallinule, black swan \\
bird & wading bird & goose, oystercatcher, little blue heron, black stork, bustard, flamingo, spoonbill \\
\midrule
container & bag & backpack, plastic bag, purse \\
container & food container & \makecell[l]{water jug, beer bottle, water bottle, wine bottle, coffee mug, vase, \\coffeepot, teapot, measuring cup, cocktail shaker} \\
\midrule
device & electronics & \makecell[l]{cassette player, cellular telephone, computer keyboard, desktop computer, \\dial telephone, hard disc, iPod, laptop} \\
device & measuring & analog clock, digital clock, wall clock, stopwatch, digital watch, odometer, barometer \\
\midrule
dog & hound & English foxhound, Italian greyhound, Afghan hound, basset, beagle, otterhound \\
dog & sporting dog & English springer, cocker spaniel, golden retriever, Irish setter \\
dog & terrier & \makecell[l]{American Staffordshire terrier, wire-haired fox terrier, standard schnauzer, \\Border terrier, Irish terrier, Yorkshire terrier} \\
dog & toy dog & papillon, Chihuahua, Japanese spaniel, Shih-Tzu, toy terrier \\
dog & working dog & \makecell[l]{collie, German shepherd, Rottweiler, miniature pinscher, \\French bulldog, Siberian husky, boxer, Eskimo dog} \\
\midrule
edible & edible fruit & Granny Smith, strawberry, lemon, orange, banana, custard apple, fig, pineapple, pomegranate \\
edible & sandwich & cheeseburger, hotdog, bagel \\
edible & vegetable & \makecell[l]{bell pepper, broccoli, cauliflower, spaghetti squash, zucchini, \\butternut squash, artichoke, cardoon, cucumber} \\
\midrule
fungus & fungus & bolete, coral fungus, earthstar, gyromitra, hen-of-the-woods, stinkhorn \\
\midrule
insect & beetle & ground beetle, ladybug, leaf beetle, long-horned beetle, tiger beetle, weevil \\
insect & butterfly & monarch, admiral, cabbage butterfly, lycaenid, ringlet, sulphur butterfly \\
insect & spider & black widow, garden spider, tarantula, wolf spider, scorpion \\
\midrule
mammal & bear & American black bear, brown bear, ice bear, sloth bear, giant panda, lesser panda \\
mammal & bovid & ox, ibex, bighorn, gazelle, impala, water buffalo, ram, bison \\
mammal & canine & \makecell[l]{Arctic fox, grey fox, red fox, African hunting dog, dingo, \\coyote, red wolf, timber wolf, white wolf, hyena} \\
mammal & equine & sorrel, zebra \\
mammal & feline & Persian cat, tabby, cheetah, jaguar, leopard, lion, snow leopard, tiger \\
mammal & great ape & chimpanzee, gorilla, orangutan \\
mammal & monkey & capuchin, spider monkey, squirrel monkey, baboon, guenon, macaque \\
\midrule
music. instr. & percussion & chime, drum, gong, maraca, marimba, steel drum \\
music. instr. & stringed & cello, violin, acoustic guitar, electric guitar, banjo \\
music. instr. & wind & bassoon, oboe, sax, flute, cornet, French horn, trombone \\
\midrule
object & ball & golf ball, ping-pong ball, rugby ball, soccer ball, tennis ball \\
object & handtool & hammer, plane, plunger, screwdriver, shovel \\
object & headdress & bathing cap, shower cap, bonnet, cowboy hat, sombrero, football helmet \\
\midrule
reptile & amphibian & bullfrog, tree frog, axolotl, spotted salamander, common newt, eft, European fire salamander \\
reptile & snake & \makecell[l]{rock python, boa constrictor, green mamba, Indian cobra, diamondback, sidewinder, \\horned viper, king snake, green snake, thunder snake} \\
reptile & turtle & box turtle, mud turtle, terrapin \\
\midrule
sea life & aqu. mammal & killer whale, grey whale, sea lion, dugong \\
sea life & bony fish & goldfish, tench, eel, anemone fish, lionfish, gar, sturgeon \\
sea-life & crab & American lobster, Dungeness crab, fiddler crab, king crab, rock crab, crayfish, hermit crab, isopod \\
sea life & shark & great white shark, tiger shark, hammerhead \\
\midrule
vehicle & bicycle & motor scooter, tricycle, unicycle, mountain bike, moped \\
vehicle & boat & speedboat, lifeboat, canoe, fireboat, gondola \\
vehicle & car & ambulance, beach wagon, cab, convertible, jeep, limousine, minivan, sports car \\
vehicle & locomotive & electric locomotive, steam locomotive \\
vehicle & sailing vessel & catamaran, trimaran, schooner \\
vehicle & truck & minivan, police van, fire engine, garbage truck, pickup, tow truck, trailer truck, school bus \\
\bottomrule
\end{tabular}

\caption{Groups and clusters of the ImageNet classes used to define the transformation queries.
}
\label{fig:clusters} 
\end{table*}


\end{document}